\documentclass[journal]{IEEEtran}

\usepackage{cite}
\usepackage{amsmath,amssymb,amsfonts}
\usepackage{algcompatible}
\usepackage{algorithm}
\usepackage{algpseudocode}  
\usepackage{graphicx}
\usepackage{textcomp}
\usepackage{xcolor,url}
\usepackage{rotating}
\usepackage{color}
\usepackage{verbatim}		
\usepackage{subfigure,balance}
\newcommand{\tabincell}[2]{\begin{tabular}{@{}#1@{}}#2\end{tabular}}
\usepackage{booktabs}
\usepackage{tikz,balance}

\usepackage{rotating}

\usepackage{xpatch}
\usepackage{multirow}
\usepackage{multicol}
\usepackage{bm}
\usepackage{epstopdf} 
\usepackage{rotating}

\usepackage{amsmath}  
\usepackage{multirow}
\usepackage{longtable}
\usepackage{tabularx}
\usepackage{supertabular}
\usepackage{adjustbox}
\usepackage{graphicx}  
\usepackage{float}  

\usepackage{color,colortbl}
\definecolor{mygrey}{gray}{.9}

\def\NoNumber#1{{\def\alglinenumber##1{}\State #1}\addtocounter{ALG@line}{-1}}

\newcommand{\D}{\mathcal{D}}

\newcommand{\M}{\mathcal{M}}

\def\NoNumber#1{{\def\alglinenumber##1{}\State #1}\addtocounter{ALG@line}{-1}}

\makeatletter
\ExplSyntaxOn
\cs_new:Npn \bibColoredItems #1#2
  {
    \clist_map_inline:nn {#2} { \cs_new:cpn {bib@colored@##1} {#1} } 
  }
\ExplSyntaxOff

\newcommand\bib@setcolor[1]{%
  \ifcsname bib@colored@#1\endcsname
    \expandafter\color\expandafter{\csname bib@colored@#1\endcsname}
  \else
    \normalcolor
  \fi
}

\xpatchcmd\@bibitem
  {\item}
  {\bib@setcolor{#1}\item}
  {}{\PatchFailed}

\xpatchcmd\@lbibitem
  {\item}
  {\bib@setcolor{#2}\item}
  {}{\PatchFailed}
\makeatother

\begin{document}

\title{Fairness-aware Multiobjective Evolutionary Learning}

\author{Qingquan Zhang,~\IEEEmembership{Member,~IEEE},
Jialin Liu,~\IEEEmembership{Senior Member,~IEEE}, Xin Yao,~\IEEEmembership{Fellow,~IEEE}

\thanks{This work has been accepted by IEEE Transactions on Evolutionary Computation.}
\thanks{Q. Zhang is with the Research Institute of Trustworthy Autonomous System, Southern University of Science and Technology, Shenzhen 518055, China. Q. Zhang is also with the Guangdong Provincial Key Laboratory of Brain-inspired Intelligent Computation, Department of Computer Science and Engineering, Southern University of Science and Technology, Shenzhen 518055, China.
J. Liu is with the Guangdong Provincial Key Laboratory of Brain-inspired Intelligent Computation, Department of Computer Science and Engineering, Southern University of Science and Technology, Shenzhen 518055, China.
J. Liu is also with the Research Institute of Trustworthy Autonomous System, Southern University of Science and Technology, Shenzhen 518055, China.
X. Yao is with the School of Data Science, Lingnan University, Hong Kong SAR.

}
\thanks{Corresponding author: Jialin Liu (liujl@sustech.edu.cn).}
    }

\maketitle

\begin{abstract}
Multiobjective evolutionary learning (MOEL) has demonstrated its advantages of training fairer machine learning models considering a predefined set of conflicting objectives, including accuracy and different fairness measures. 
Recent works propose to construct a representative subset of fairness measures as optimisation objectives of MOEL throughout model training. 
However, the determination of a representative measure set relies on dataset, prior knowledge and requires substantial computational costs.
What's more, those representative measures may differ across different model training processes. 
Instead of using a static predefined set determined before model training, this paper proposes to dynamically and adaptively determine a representative measure set online during model training. The dynamically determined representative set is then used as optimising objectives of the MOEL framework and can vary with time.
Extensive experimental results on 12 well-known benchmark datasets demonstrate that our proposed framework achieves outstanding performance compared to state-of-the-art approaches for mitigating unfairness in terms of accuracy as well as 25 fairness measures although only a few of them were dynamically selected and used as optimisation objectives. The results indicate the importance of setting optimisation objectives dynamically during training. 
\end{abstract}


\IEEEpeerreviewmaketitle

\section{Introduction}\label{sec:intro} 

\IEEEPARstart{F}{airness} is a critical concern in artificial intelligence~\cite{huang2022overview,pessach2022review, yu2022towards,mehrabi2021survey}. Over the years, at least 20 different measures to quantify (un)fairness have been proposed~\cite{hutchinson201950}. Different fairness measures often exhibit complex relationships among them~\cite{pessach2022review, yu2022towards,mehrabi2021survey,hutchinson201950, garg2020fairness,fairenough2023,yuan2024FairerML}, such as conflicts, inconsistencies or even unknown patterns. Additionally, fairness is often conflicting with the accuracy of learning models~\cite{pessach2022review, yu2022towards,mehrabi2021survey}. Thus, the intricate relationships among accuracy and multiple fairness measures pose great challenges in fair machine learning and fair artificial intelligence in general.

Various techniques have been developed to optimise fairness measures of learning models, which can be mainly divided into two categories~\cite{pessach2022review, yu2022towards,mehrabi2021survey}. The core idea of the first category converts accuracy and multiple fairness measures into one combined objective to be optimised. One such technique is Multi-FR~\cite{wu2022multifr}, which calculates the losses of all the measures and uses a weighted sum to update a learning model. Multiobjective evolutionary learning (MOEL)~\cite{chandra2006ensemble}, as the second category, demonstrates significant advantages in training fairer machine learning models~\cite{QingquanFair2021,Qingquan2022Ensemble,gsh2023fairer,haas2019price,la2023optimizing}. In this category, a learning algorithm operates by explicitly defining a set of measures, including accuracy and multiple fairness measures, and simultaneously optimising these measures during the training process, where each measure is viewed as an objective~\cite{pessach2022review, yu2022towards,mehrabi2021survey}. MOEL can generate a diverse set of learning models, with each model representing a tradeoff among different measures. 

Optimising all the fairness measures may not always be necessary. Recent studies~\cite{anahideh2021choice,fairenough2023,mehrabi2021survey} show that a comprehensive set of fairness measures can be represented by a subset because of the positive correlations among some measures. Such a subset is denoted as a \textit{representative measure subset}. 
In light of this, the work of~\cite{Qingquan2022Ensemble} proposed to optimise a representative subset of measures~\cite{anahideh2021choice} throughout the model training. The findings of~\cite{Qingquan2022Ensemble} show that by optimising this carefully selected subset, improvements can be achieved across all the fairness measures, even including those that were not used as optimisation objectives during model training.

However, using a static predefined measure subset~\cite{Qingquan2022Ensemble} still has limitations for three reasons. First, prior knowledge or significant computational costs were involved in finding a subset that can comprehensively represent all the measures~\cite{fairenough2023,anahideh2021choice}. 
Second, the proper representative subset may vary across different datasets. 
Carefully determining the representative subset of measures for specific datasets requires extra computational cost before model training.
Third, the correlation among accuracy and multiple fairness measures is changing along with the model training stages. The optimal representative subset of measures at one optimisation stage may not necessarily remain optimal at another stage.

\begin{figure}[htbp]
  \begin{center}  \includegraphics[width=0.48\textwidth]{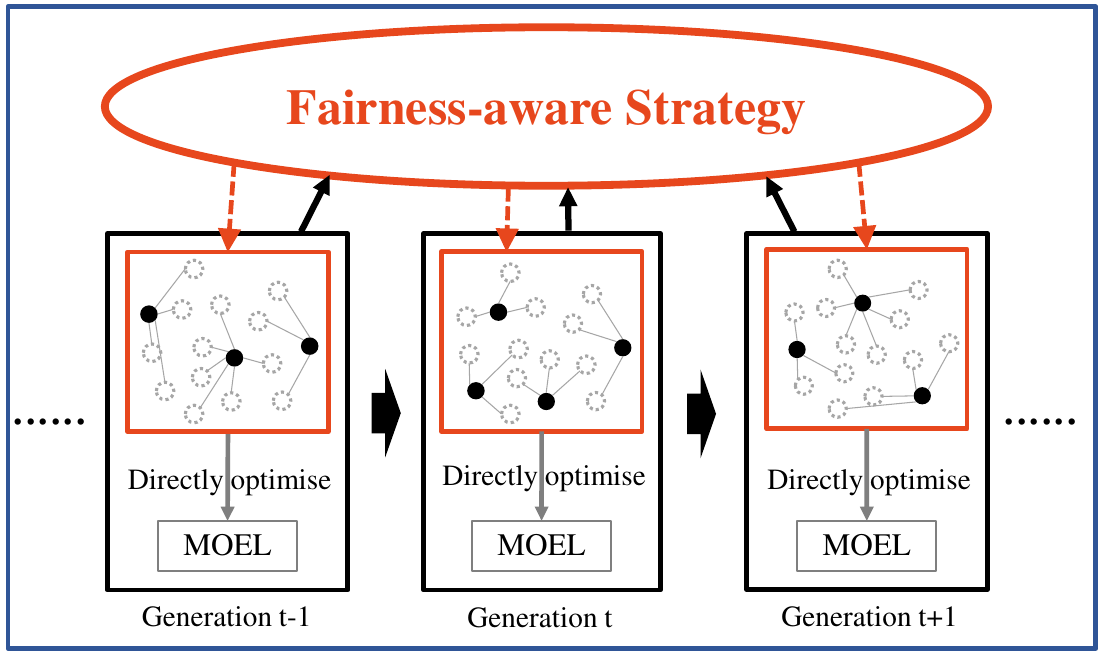}
  \end{center}
  \caption {Flow of our framework, where a fairness-aware strategy is used to dynamically select a representative subset (solid circles) to be optimised using MOEL at each generation to improve all the measures (all circles). Each generation involves mating selection, reproduction and survival selection, encompassing the loop outlined in lines 6-13 of Algorithm 1, which will be detailed in Section III.A.
  } \label{fig:process_overview}
\end{figure}

Because of the aforementioned issues, instead of using a static predefined one, an adaptively online-determined representative subset that does not need any prior knowledge is more promising as the optimising objectives during model training. 
Novel contributions of this paper are as follows:
\begin{enumerate}
        \item We introduce a 
        \textbf{F}airness-\textbf{a}ware strategy using \textbf{M}ulti\textbf{O}bjective \textbf{E}volutionary \textbf{L}earning (FaMOEL) framework to optimise an objective set including accuracy and multiple fairness measures, as shown in Fig. \ref{fig:process_overview}. 
        Along with the training process of FaMOEL, our framework is aware of the current model training process and then adaptively determines a representative subset (solid circles in Fig. \ref{fig:process_overview}) of measures as objectives instead of using a predefined static one. Thus, the representative subset may vary along with the model training process.

        \item Based on our framework, an efficient instantiation algorithm is developed. Specifically, we design and incorporate three enhancement strategies into ORNCIE~\cite{wang2016objective}, aiming to construct the most appropriate representative measure subset.  
\end{enumerate}

We demonstrate the effectiveness of our framework and its instantiation on 12 benchmark datasets. Empirical results observed by four performance indicators reveal that our framework achieves outstanding performance in terms of accuracy and 25 fairness measures compared to the state-of-the-art methods\footnote{Code of this work is available at \url{https://github.com/qingquan63/FaMOEL}}. Our framework can well perceive a suitable representative subset according to the current model training process. The results also demonstrate that the most appropriate representative subset did vary from generation to generation.

The remainder of this paper is organised as follows. Section \ref{sec:background} introduces the background. Our proposed framework FaMOEL and the designed algorithm based on this framework are presented in Section \ref{sec:framework}.
Section \ref{sec:exp} gives the experimental results.
Section \ref{sec:conclusion} concludes the paper and discusses future work.

\section{Background}\label{sec:background}
This section presents an overview of fairness measures and their relationship. Then, the multiobjective evolutionary learning framework for fairer machine learning is introduced.

\subsection{Fairness Measures in Machine Learning}
Numerous measures have been proposed to evaluate (un)fairness from ethical standpoints in the context of fairness~\cite{pessach2022review, yu2022towards,mehrabi2021survey,hutchinson201950, garg2020fairness,fairenough2023,yuan2024FairerML}. There is no consensus on a universally agreed fairness measure that is capable of comprehensively taking into account all perspectives of fairness. Many measures exhibit diverse and intricate relationships with one another. Some measures demonstrate positive correlations with others, while some others present conflicts. Additionally, accuracy and some fairness measures are often conflicting or inconsistent with each other~\cite{hutchinson201950, fairenough2023}. 

Based on the confusion matrix (shown in Table \ref{tab:confusion}) or other principles, the study of~\cite{fairenough2023} comprehensively reviews 25 fairness measures, which are detailed in Table \ref{tab:Fairness25}. Given a sensitive attribute (i.e., gender, race), a dataset can be divided into two groups, unprivileged group $g_u$ and privileged group $g_p$. The fairness measures Fair1-Fair13 are formulated based on the confusion matrix  of groups $g_u$ and $g_p$. In Fair6 and Fair11, $ERR$ is equal to $(FN+FP)/(TP+FP+FN+TN)$. $G$, $y$ and $\hat{y}$ denote the sensitive attributes, true labels and predicted labels obtained by a learning model, respectively. In Fair16-Fair24, $|G|$ is the number of groups, $n_g$ refers to the size of group $g$, $n$ is the number of observations (i.e., $n=\sum_{g} n_g$), and $\alpha$ is a positive constant. The benefit vector $b_i$ is equal to $\hat{y_i}-y_i+1$ for the $i$-th data. $b^{g_u}_i$ and $b^{g_d}_i$ are the benefit values of the $i$-th data in unprivileged and privileged groups, respectively. $\mu$, $\mu_{g_u}$ and $\mu_{g_d}$ are the mean values of all the $b_i$, $b^{g_u}_i$ and $b^{g_d}_i$, respectively.

The comprehensive analysis provides insights into the relationships and tradeoffs among these 25 fairness measures~\cite{fairenough2023}. It suggests that it is possible to cluster all those fairness measures into six groups based on their correlation~\cite{fairenough2023}. Specifically, the conflicts and consistencies among these measures are analysed through a number of trained models obtained by three algorithms, including the logistic regression, a reweighing method~\cite{kamiran2012data} and a meta fair method~\cite{celis2019classification}, across seven datasets. Then, the obtained six groups can help to construct a representative measure subset. Note that this process needs significant computational cost~\cite{fairenough2023}.

However, some fairness measures are positively correlated in one dataset, while they are negatively correlated in other datasets~\cite{garg2020fairness,anahideh2021choice}. Therefore, a proper representative subset for one dataset may not be suitable for other datasets.

\begin{table}[htbp]

\caption{Confusion matrix}
\centering
\setlength{\tabcolsep}{3mm}{
\begin{tabular}{cccll}

\cline{1-3}
\multicolumn{1}{|c|}{} & \multicolumn{1}{c|}{Actual positive} & \multicolumn{1}{c|}{Actual negative}                                                                            &  &  \\ \cline{1-3}
\multicolumn{1}{|c|}{Predicted   positive} & \multicolumn{1}{c|}{\begin{tabular}[c]{@{}c@{}}TP\\      PPV = TP/(TP+FP)\\      TPR = TP/(TP+FN)\end{tabular}} & 

\multicolumn{1}{c|}{\begin{tabular}[c]{@{}c@{}}FP\\      FDR = FP/(TP+FP)\\      FPR = FP/(FP+TN)\end{tabular}} &  &  \\ \cline{1-3}

\multicolumn{1}{|c|}{Predicted   negative} & \multicolumn{1}{c|}{\begin{tabular}[c]{@{}c@{}}FN\\      FOR = FN/(TN+FN)\\      FNR = FN/(TP+FN)\end{tabular}} & 

\multicolumn{1}{c|}{\begin{tabular}[c]{@{}c@{}}TN\\      NPV = TN/(TN+FN)\\      TNR = TN/(TN+FP)\end{tabular}} &  &  \\ \cline{1-3}

\multicolumn{1}{l}{} & \multicolumn{1}{l}{} & \multicolumn{1}{l}{} &  & 
\end{tabular}
}
\label{tab:confusion}
\end{table}

\begin{table*}[htbp]
  \centering
  \caption{Summary of 25 fairness measures~\cite{fairenough2023}}
    \begin{tabular}{ccc}
    \toprule
    Notation & Name & \multicolumn{1}{c}{Formulation} \\
    \midrule
    Fair1    & True positive rate difference &  $TPR(g_u) -TPR(g_p)$\\
    Fair2    & False positive rate difference & $FPR(g_u) - FPR(g_p)$ \\
    Fair3    & False negative rate difference & $FNR(g_u) - FNR(g_p)$\\
    Fair4    & False omission rate difference & $FOR(g_u) - FOR(g_p)$\\
    Fair5    & False discovery rate difference & $FDR(g_u) - FDR(g_p)$\\
    Fair6   & Error rate difference & $ERR(g_u) - ERR(g_p)$ \\
    Fair7    & False positive rate ratio & $FPR(g_u) / FPR(g_p)$ \\
    Fair8    & False negative rate ratio & $FNR(g_u) / FNR(g_p)$ \\
    Fair9    & False omission rate ratio & $FOR(g_u) / FOR(g_p)$ \\
    Fair10    & False discovery rate ratio & $FDR(g_u) / FDR(g_p)$ \\
    Fair11   & Error rate ratio &  $ERR(g_u) / ERR(g_p)$ \\
    Fair12   & Average odds difference &  $\frac{1}{2} (TPR(g_u) - TPR(g_p) + FPR(g_u) - FPR(g_p))$\\
    Fair13   & Average abs odds difference & $\frac{1}{2}$ ($|TPR(g_u) - TPR(g_p)| + |FPR(g_u) - FPR(g_p)|)$\\
    
    Fair14   & Disparate impact & $P(\hat{y}=1 | G=g_u) / P(\hat{y}=1 | G=g_p) $ \\
    Fair15   & Statistical parity difference & $P(\hat{y}=1 | G=g_u) - P(\hat{y}=1 | G=g_p) $ \\
    Fair16   & Generalized entropy index &  $\frac{1}{n\alpha(\alpha-1)} \sum_{i=1}^{n} \left[\left(\frac{b_i}{\mu}\right)^\alpha-1\right] $ \\
    Fair17   & Between all groups generalized entropy index &  $\frac{1}{n\alpha(\alpha-1)} \sum_{g=1}^{|G|} n_g\left[ \left(\frac{\mu_{g}}{\mu}\right)^{\alpha}-1 \right]$ \\
    Fair18   & Between group generalized entropy index & $\frac{1}{n\alpha(\alpha-1)} n_{g_u}\left[ \left(\frac{\mu_{g_u}}{\mu}\right)^{\alpha}-1 \right]$ + $\frac{1}{n\alpha(\alpha-1)} n_{g_d}\left[ \left(\frac{\mu_{g_d}}{\mu}\right)^{\alpha}-1 \right]$ \\
    
    Fair19   & Theil index &  $\frac{1}{n} \sum_{i=1}^{n} \frac{b_i}{\mu}ln\frac{b_i}{\mu} $ \\
    
    Fair20   & Coefficient of variation &  $2\sqrt{\frac{1}{n} \sum_{i=1}^{n} \frac{b_i}{\mu}ln\frac{b_i}{\mu}}$ \\
    
    Fair21   & Between group theil index &  $\frac{1}{n_{g_u}} \sum_{i=1}^{n_{g_u}} \frac{b^{g_u}_i}{\mu_{g_u}}ln\frac{b^{g_u}_i}{\mu_{g_u}} + \frac{1}{n_{g_p}} \sum_{i=1}^{n_{g_p}} \frac{b^{g_p}_i}{\mu_{g_p}}ln\frac{b^{g_p}_i}{\mu_{g_p}}$  \\
    
    Fair22   & Between group coefficient of variation & $2\sqrt{\frac{1}{n_{g_u}} \sum_{i=1}^{n_{g_u}} \frac{b^{g_u}_i}{\mu_{g_u}}ln\frac{b^{g_u}_i}{\mu_{g_u}}} + 2\sqrt{\frac{1}{n_{g_p}} \sum_{i=1}^{n_{g_p}} \frac{b^{g_p}_i}{\mu_{g_p}}ln\frac{b^{g_p}_i}{\mu_{g_p}}}$ \\
    
    Fair23   & Between all groups theil index & $\sum_{g=1}^{|G|} \frac{1}{n_g} \sum_{i=1}^{n_g} \frac{b^g_i}{\mu_g}ln\frac{b^g_i}{\mu_g}$  \\
    
    Fair24   & Between all groups coefficient of variation &  $\sum_{g=1}^{|G|} 2\sqrt{\frac{1}{n_g} \sum_{i=1}^{n_g} \frac{b^g_i}{\mu_g}ln\frac{b^g_i}{\mu_g}}$\\
    
    Fair25   & Differential fairness bias amplification & \tabincell{c}{Difference in smoothed empirical differential fairness \\ between the classifier and the original dataset~\cite{foulds2020intersectional}} \\
    \bottomrule
    \end{tabular}%
  \label{tab:Fairness25}%
\end{table*}%

\subsection{Mitigating Unfairness through Multiobjective Evolutionary Learning}\label{sec:MOEL}

Multiobjective evolutionary learning (MOEL)~\cite{chandra2006ensemble} has been proposed to optimise accuracy and multiple fairness measures for fairer machine learning~\cite{QingquanFair2021,Qingquan2022Ensemble,gsh2023fairer,yuan2024FairerML}, aiming to evolve a population of learning models, e.g., artificial neural nets (ANNs), by utilising multiobjective evolutionary algorithms (MOEAs)~\cite{li2015many}. MOEAs represent a class of optimisation techniques specifically designed to address problems with multiple conflicting objectives. Unlike traditional single-objective optimisation methods, a set of optimal solutions, called optimal Pareto front is desired when solving multiobjective optimisation problems~\cite{li2019quality}. MOEAs typically aim to approximate the optimal Pareto front by maintaining a set of solutions. When evaluating the solution set, four aspects are considered, namely convergence, spread, uniformity, and cardinality~\cite{li2019quality}.

In mitigating unfairness, MOEL~\cite{QingquanFair2021,Qingquan2022Ensemble,gsh2023fairer,yuan2024FairerML} can provide a diverse model set, where a model in the set indicates a tradeoff among the accuracy and different fairness measures. Later, study~\cite{Qingquan2022Ensemble} constructs an ensemble model with the diverse model set to automatically balance accuracy and fairness measures. In MOEL-based algorithms~\cite{QingquanFair2021,Qingquan2022Ensemble}, it's worth noting that the objectives considered, which include accuracy and fairness measures, may not always be differentiable. Furthermore, study~\cite{gsh2023fairer} confirms that leveraging an appropriate gradient from adversarial learning can significantly enhance model fairness when performing partial training~\cite{yao1997new,yao1999evolving}. However, this method is limited to optimising only two fairness measures: equalised odds and demographic parity.

Given a set of fairness measures, it is not always necessary to optimise all of them thanks to positive correlations among some measures~\cite{ anahideh2021choice,mehrabi2021survey,fairenough2023}. Particularly, the studies of~\cite{ anahideh2021choice} analyse the relationship among many fairness measures and select representative measures.
Motivated by this observation, the study of~\cite{Qingquan2022Ensemble} uses accuracy and the selected representative fairness measures~\cite{anahideh2021choice} as the optimisation objectives in MOEL. Note that throughout the entire model training process, even on different datasets, the study~\cite{Qingquan2022Ensemble} only optimises those predefined measures. 

However, as mentioned before, each dataset may have a different suitable representative fairness subset~\cite{garg2020fairness,anahideh2021choice}. 
Determining a suitable representative fairness subset for a new dataset consumes significant computational cost.
Furthermore, even when dealing with the same dataset in one model training trial, the suitable representative subset may vary across different training stages.

\section{Fairness-aware Multiobjective Evolutionary Learning}\label{sec:framework} 
Section \ref{sec:FaMOEL} presents our proposed framework, namely FaMOEL, which dynamically and adaptively determines a representative subset of fairness measures during model training without any prior knowledge. The determined set are used as objectives of MOEL to guide the evolution of learning models. Then, an instantiation algorithm based on FaMOEL is implemented in Section \ref{sec:FaMOEL_algo}, where three enhancement strategies are designed to improve the fairness-awareness ability of our method.

\subsection{Fairness-aware Multiobjective Evolutionary Learning Framework for Mitigating Unfairness}\label{sec:FaMOEL}

First, in our study, we formulate the task of improving the learning model's accuracy and fairness as a multi-objective learning task~\cite{QingquanFair2021,Qingquan2022Ensemble,gsh2023fairer,yuan2024FairerML}
\begin{equation}
    minimise_{\mathbf{x} \in \Omega} \quad F(\mathbf{x}) = \{f_1(\mathbf{x}), f_2(\mathbf{x}),\dots,f_M(\mathbf{x})\},
\end{equation}
where $\mathbf{x}$ represents the parameters of a learning model within the decision space $\Omega$. $F(\mathbf{x})$ is a set of $M$ objective functions  that assess the accuracy and fairness of the model parametrised by $\mathbf{x}$ on the given task.

Algorithm \ref{algo:framework} outlines our proposed FaMOEL with six inputs, including an initial population of models $\mathcal{M}$, a set of model evaluation objectives $\mathcal{E}$, a fairness-aware strategy $FA$, training data $\mathcal{D}_{train}$, validation data $\mathcal{D}_{validation}$, and a multiobjective optimiser $\pi$. The objectives in $\mathcal{E}$ are used to calculate optimised objective values, such as accuracy and fairness measures, based on the predictions of the models in $\mathcal{M}$ on the validation data $\mathcal{D}_{validation}$. Fairness-aware strategy $FA$ is used to find a representative subset from the entire objectives $\mathcal{E}$ during model training. Note that compared with the previous work~\cite{QingquanFair2021}, as clearly illustrated in Fig. \ref{fig:process_overview}, the core difference is the fairness-aware strategy $FA$. The training data $\mathcal{D}_{train}$ is utilised for local search strategies, such as partial training~\cite{yao1997new,yao1999evolving}, to update the parameters of the models in $\mathcal{M}$.

\begin{algorithm}[htbp]
\caption{\label{algo:framework}Fairness-aware multiobjective learning framework.}
\begin{algorithmic}[1]
\Require Initial models $\M_1,\dots,\M_\lambda$, set of model evaluation objectives $\mathcal{E}$, training dataset $\mathcal{D}_{train}$, validation dataset $\mathcal{D}_{validation}$, multiobjective optimiser $\pi$, fairness-aware strategy $FA$
\Ensure A final model set $\M_1,\dots,\M_\lambda$
\State Partially train~\cite{yao1997new,yao1999evolving} $\M_1,\dots,\M_\lambda$ over $\mathcal{D}_{train}$ \label{line:partial1}
\For{$i \in \{1,\dots,\lambda\}$}
    \State{${\epsilon}_i \leftarrow$ Evaluate ${\M}_i$ with objectives $\mathcal{E}$ on $\D_{validation}$}
\EndFor
\While{terminal conditions are not fulfilled}
    \State $\mathcal{E'} \leftarrow$ Perform fairness-aware $FA$ to select \NoNumber{representative objectives from $\mathcal{E}$} \label{line:fairness_aware}
    \State $\mathcal{P} \leftarrow$ Select $\mu$ promising models from $\M_1,\dots,\M_\lambda$ \NoNumber{with ``best''  $\epsilon_1,\dots,\epsilon_\mu$ according to $\pi$ and $\mathcal{E'}$} \label{line:parentselection}
    \State $\M' \leftarrow$ Generate $\phi$ new models $\M_1',\dots, \M_\phi'$ from $\mathcal{P}$ \NoNumber{according to $\pi$} \label{line:generateoff}
\For{$i \in \{1,\dots,\phi\}$}
    \State{${\M'}_i\leftarrow$ Partially train~\cite{yao1997new,yao1999evolving} $\M'_i$ on $\D_{train}$} \label{line:partial2}
    \State{${\epsilon'}_i \leftarrow$ Evaluate ${\M'}_i$ with objectives $\mathcal{E}$ on $\D_{validation}$}
\EndFor
    \State{$<\M_1,\epsilon_1>,\dots, <\M_\lambda,\epsilon_\lambda> \leftarrow$ Select $\lambda$ promising \NoNumber{models from $\{\M_1,\dots,\M_\lambda\} \bigcup \{\M_1',\dots, \M_\phi'\}$} by $\pi$ \NoNumber{ and $\mathcal{E'}$} based on $\epsilon_1,\dots,\epsilon_\lambda$ and $\epsilon'_1,\dots,\epsilon'_\phi$, and } then \NoNumber{update $\M_1,\dots,\M_\lambda$ and $\epsilon_1,\dots,\epsilon_\lambda$ accordingly} \label{line:envselection}
\EndWhile
\end{algorithmic}
\end{algorithm}

The multiobjective optimiser $\pi$ consists of three main strategies~\cite{QingquanFair2021,Qingquan2022Ensemble}: reproduction, mating selection, and survival selection.
In our framework, during model initialisation and generation, partial training~\cite{yao1997new,yao1999evolving} is always applied to the models using the training data $\mathcal{D}_{train}$. The objective values of each model are obtained through the evaluation objectives $\mathcal{E}$ (lines \ref{line:partial1} and \ref{line:partial2} in Algorithm \ref{algo:framework}). In the main loop, the fairness-aware strategy $FA$ is performed to online select a representative subset $\mathcal{E'}$ from the evaluation objectives $\mathcal{E}$ according to the current evolution process (line \ref{line:fairness_aware} in Algorithm \ref{algo:framework}). Then, the mating selection strategy of $\pi$ selects a promising set of parent models $\mathcal{P}$ from the population $\mathcal{M}$ (line \ref{line:parentselection} in Algorithm \ref{algo:framework}) only considering $\mathcal{E'}$. After that, $\phi$ new models, denoted as $\mathcal{M}'$, are generated by inheriting information from $\mathcal{P}$ through the reproduction strategy of $\pi$ (line \ref{line:generateoff} in Algorithm \ref{algo:framework}). This strategy modifies the parameters of the parent models, often using operators like crossover and mutation. After partial training and model evaluation, $\lambda$ candidate models are selected from the combination of the original population $\mathcal{M}$ and the new models $\mathcal{M}'$ using the survival selection strategy of $\pi$ considering the representative objectives $\mathcal{E'}$ (line \ref{line:envselection} in Algorithm \ref{algo:framework}). These selected models form the updated population $\mathcal{M}$ for the next generation. These steps are repeated until a termination criterion is met.

Finally, a model set $\M_1,\dots,\M_\lambda$ is obtained, from which decision makers can select one or multiple to deploy according to specific requirements in real-world scenarios.

\subsection{Instantiation Algorithm based on Our Framework}\label{sec:FaMOEL_algo}
To verify the effectiveness of our framework FaMOEL, an instantiation algorithm based on FaMOEL is developed and the key components of FaMOEL are introduced as follows, including the model set, evaluation objectives, fairness-aware method and multiobjective optimisation algorithm. Noted that our proposed framework allows for flexibility in selecting these components based on specific prediction tasks and preferences.

\subsubsection{Model Set} A range of machine learning (ML) models can be utilised within our framework. In our study, a collection of artificial neural networks (ANNs) with the same architecture is employed as individuals. Each ANN's weights and biases are encoded as a real-value vector and represented as an individual~\cite{yao1997new,Qingquan2022Ensemble}.

\subsubsection{Evaluation objectives} 
In this study, a total of 26 fairness measures, including accuracy and Fair1 to Fair25 (as listed in Table \ref{tab:Fairness25}), are considered. The accuracy is evaluated using the cross-entropy ($CE$) measure commonly employed for classifiers~\cite{Qingquan2022Ensemble}, and it is minimised.
Following~\cite{Qingquan2022Ensemble}, the absolute values of Fair1--Fair6, Fair12, Fair13 and Fair15 are minimised.
For Fair7--Fair11 and Fair14 using ratios, we construct the objective functions to be minimised with the transformation following the work of~\cite{Qingquan2022Ensemble}. Taking Fair7 ($\frac{FPR(g_u)}{FPR(g_p)}$) as an example, its corresponding objective function is calculated as $1 - \min{ \{\frac{FPR(g_u)}{FPR(g_p)}, \frac{FPR(g_p)}{FPR(g_u)}} \}$.
Fair16--Fair25 are directly used as objective values to be minimised since their values are always positive.
The transformed objectives corresponding to Fair1--Fair25 are denoted as $f_1$--$f_{25}$, respectively.
The optimal values of $f_1$--$f_{25}$ are all zeros.

\subsubsection{Multiobjective Optimiser}
Parent selection, survival selection and reproduction strategy of $\pi$ can be implemented by any multiobjective evolutionary algorithm.

In our instantiation algorithm, we utilise Two\_Arch2~\cite{wang2015twoarch2} for both parent selection and survival selection. The survey~\cite{li2023multiobjective} demonstrates the efficacy of Two\_Arch2 to address many-objective optimisation.
Two\_Arch2 is popular and efficient~\cite{li2015many,li2023multiobjective,9819828,liu2022multiobjective}, exhibiting competitive performance in handling many-objective optimisation problems, which maintains two archives, each focusing on convergence and diversity of individuals, respectively.

In the reproduction strategy, isotropic Gaussian perturbation and the variant of weight crossover are applied as mutation and crossover operators~\cite{Qingquan2022Ensemble}, respectively. A set of new $\M'$ (line \ref{line:generateoff} in Algorithm \ref{algo:framework}) can be obtained by applying the mutation operator to the model set resulting from the crossover between the convergence archive and the diversity archive~\cite{wang2015twoarch2}. 

Specifically, isotropic Gaussian perturbation~\cite{Qingquan2022Ensemble} is performed as the mutation operator, formulated as
$r_i = r_i + \delta_i$, 
where the $i$-th weight of an ANN, denoted as $r_i$, undergoes isotropic Gaussian perturbation, with $\delta_i$ sampled from a normal distribution $\mathcal{N}(0,\sigma^2)$. $\sigma$ represents the mutation strength. Given parents $p$ and $q$, the weight crossover~\cite{Qingquan2022Ensemble} is applied as

    $r_i^{o_1} =u_i r_i^p  +  (1 - u_i)r_i^q$, 
    $r_i^{o_2} = u_i r_i^q  +  (1 - u_i)r_i^p$,
where $u_i$ is sampled from (0,1) uniformaly at random. Meanwhile, $r_i^{p}$, $r_i^{q}$, $r_i^{o_1}$, and $r_i^{o_2}$ denote the $i$-th weight of the parent $p$, parent $q$, offspring $o_1$, and offspring $o_2$, respectively.

Regarding the partial training~\cite{yao1997new,yao1999evolving, Qingquan2022Ensemble}, model parameters are updated by the stochastic gradient descent (SGD) optimiser~\cite{ruder2016overview}.

\subsubsection{Fairness-aware Strategy}\label{sec:Fairness_aware_strategy}

The fairness-aware strategy is to adaptively construct a representative subset from all the objectives to be considered according to the current evolution status of each generation. 
In the literature, objective subset selection methods can be generally divided into three categories~\cite{li2023offline}: dominance-based, model-based and correlation-based methods. 
Dominance-based methods~\cite{5699918,yuan2017objective} aim to identify representative objectives that preserve the dominance structure as much as possible. In contrast, model-based methods~\cite{li2019hyperplane,li2023offline} build a model to approximate the obtained non-dominated front and select representative objectives based on the model's coefficients. However, both dominance-based and model-based methods are less effective for problems with many objectives (up to 15)~\cite{li2023offline}. 
Dominance-based methods struggle to maintain the informative dominance structure when there is a high proportion of non-dominated solutions~\cite{yuan2017objective}. The models constructed by model-based methods often lack accuracy due to the relatively limited number of non-dominated solutions in a highly approximated space with many objectives~\cite{li2023offline}. In contrast, correlation-based methods~\cite{6151114,wang2016objective} select representative objectives by leveraging the correlation relationships among objectives. Among these, ORNCIEE~\cite{wang2016objective} has proven to be effective even in problems with up to 50 objectives. Therefore, we have chosen ORNCIE as the fairness-aware strategy in our study.

Inspired by ORNCIE~\cite{wang2016objective}, we propose our fairness-aware strategy, shown in Algorithm \ref{algo:reduction}, by adding three novel enhancement strategies based on ORNCIE. ORNCIE calculates a modified nonlinear correlation information entropy (mNCIE)~\cite{wang2016objective} to analyse the interrelationships among objectives according to the current population information and subsequently identify a representative subset. The obtained subset is directly optimised by a multiobjective optimiser, such as Two\_Arch2~\cite{wang2015twoarch2}.

\begin{algorithm}[htbp]
\caption{\label{algo:reduction}Fairness-aware strategy.}
\begin{algorithmic}[1]
\Require Current generation $t$, set of model evaluation objectives $\mathcal{E}$, history of mNCIE matrices $NC = \{NC_1, \dots, NC_{t-1} \}$, objective values of current population $\epsilon$, selection threshold $\tau$
\Ensure Set of representative objectives $\mathcal{E'}$, history of mNCIE matrices $NC$

\State $NC_t \leftarrow$ Calculate the mNCIE matrix~\cite{wang2016objective} according to the objective values of current population $\epsilon$ \label{line:calmNICE}

\State $NC = NC \bigcup \{NC_t\}$ \label{line:addmNICE}
 
\If {$t < 10$} \label{algo:warm}
    \State $\mathcal{E'} = \mathcal{E}$ \label{line:dir_ouput}
\Else {} 
    \State $NC^r \leftarrow$ Calculate a mNCIE matrix according \NoNumber{to the matrices $NC$ of the last 10 generations} \label{line:NCIEr}
    \State $S = [1, 2, \dots, |\mathcal{E}| ]$
    \State $\mathcal{E'} = \emptyset$
    
    \While{$S \neq \emptyset$}
        \If {all the elements in $NC^r$ are positive}
            \State $J=argmax_j(sum(NC^r(1:|\mathcal{E}|, j)))$, where  \NoNumber{ $j \in S$}
        \Else {} 
            \State $J=argmin_j(sum(NC^r(i, j)))$,  where  \NoNumber{ $NC^r(i,j) < 0$, $1\leq i \leq m$ and $j \in S$}
        \EndIf
        \State $S = S / \{ J\}$
        \State$\mathcal{E'} = \mathcal{E'} \bigcup \{ \mathcal{E}_J \}$
        \State $Del = \{j| NC^r(J, j) > \tau \}$, where $j \in S$ \label{line:cut}
        \State $S = S_t / Del$
    \EndWhile
\EndIf
\end{algorithmic}
\end{algorithm}

\begin{table*}[h]
  \centering
  \caption{12 benchmark datasets used in our study}
  \begin{adjustbox} {max width=\linewidth}
  \setlength{\tabcolsep}{0.01mm}{
    \begin{tabular}{ccccccc}
    \toprule
    Dataset & Source & Domain & Sensitive & \tabincell{c}{\{Privileged, \\ Unprivileged\}} & Description of Prediction Task & \tabincell{c}{Imbalance \\ Rate(+:-)}  \\
    \midrule
    Heart health & \tabincell{c}{Faisalabad Institute of Cardiology \\ and  Allied Hospital in Faisalabad}  & Healthcare & Age   & \{Young, Old\}& \tabincell{c}{Whether a person will have heart disease or not} &1.17:1\\
    
    Titanic & Titanic disaster passenger list & Disaster & Gender & \{Male, Female\} & \tabincell{c}{Whether a person will survive sinking of Titanic or not} &1:1.61\\
    
    German &  - & Finance & Gender & \{Male, Female\} & \tabincell{c}{Whether a person has an acceptable credit risk or not} &2.33:1\\
    
    Student performance & Two Portuguese secondary schools & Education & Gender & \{Male, Female\} & \tabincell{c}{Whether a student will pass the exam or not} &1:1.57\\
    
    COMPAS & \tabincell{c}{ The Broward County record} &Criminology & Gender & \{Male, Female\} & \tabincell{c}{Whether an arrested offender will be rearrested \\within two years counting from taking the test or not} &1:1.20\\
    
    Bank & \tabincell{c}{A Portuguese banking institution\\ direct marketing campaigns} & Finance  & Age   & \{Old, Young\} & \tabincell{c}{Whether a client will subscribe to a term or not} & 1:7.55\\
    
    Adult & Census bureau database & Finance & Gender & \{Male, Female\} & \tabincell{c}{Whether a person can get income higher or not} &1:3.03\\
    
    Drug consumption & \tabincell{c}{Rampton Hospital online survey } & Healthcare & Gender & \{Female, Male\} & \tabincell{c}{Whether a person never used mushroom before or not} &1.72:1\\
    
    Patient treatment & \tabincell{c}{An Indonesia private hospital} & Healthcare  & Gender & \{Male, Female\}& \tabincell{c}{Whether a patient will be in care or not} &1.52:1\\
    
    LSAT & \tabincell{c}{An American Law School Admission\\ Council survey across 163 law schools} & Education  & Gender & \{Male, Female\} & \tabincell{c}{Whether a student will pass the exam or not} &8.07:1\\
    
    Default & \tabincell{c}{Chinese (Taiwan) customers' \\ default payments} & Finance & Gender & \{Male, Female\} & \tabincell{c}{Whether a customer will default on payments or not} &1:7.55\\
    
    Dutch & \tabincell{c}{Dutch aggregated groups }  & Finance & Gender & \{Male, Female\} & \tabincell{c}{Whether a person has a highly prestigious or not} &1:1.10\\
    \bottomrule
    \end{tabular}%
    }
  \label{tab:dataset}%
  \end{adjustbox}
\end{table*}%

As described in Algorithm \ref{algo:reduction}, our proposed fairness-aware strategy takes five inputs, including current generation $t$, set of model evaluation objectives $\mathcal{E}$, history of mNCIE matrices $NC = \{NC_1, \dots, NC_{t-1} \}$, selection threshold $\tau$, objective values of current population $\epsilon$ (obtained from Algorithm \ref{algo:framework}).
Firstly, mNCIE $NC_t$~\cite{wang2016objective} is calculated according to $\epsilon$ (line \ref{line:calmNICE} in Algorithm \ref{algo:reduction}). 
$NC_1$ is a symmetry matrix with a size of $|\mathcal{E}| \times |\mathcal{E}|$, where each value in the matrix falls within the range of $[-1,1]$. A large positive value between a pair of objectives indicates that the two objectives are highly positively correlated. On the contrary, a low negative value between a pair of objectives suggests a strong negative correlation between them.

Then, $NC_t$ is appended to the historical mNCIE matrices (line \ref{line:addmNICE} in Algorithm \ref{algo:reduction}). If the generation number $t$ is less than 10, the entire objective set $\mathcal{E}$ is used as objectives to be optimised (line \ref{line:dir_ouput} in Algorithm \ref{algo:reduction}), which is viewed as warm starting. The warm starting aims to enhance the exploration ability of the evolution in the early stage by considering the entire $\mathcal{E}$, which is the first enhancement. After 10 generations, $NC_r$ is calculated by taking into account the last 10 matrices in $NC$ (line \ref{line:NCIEr} in Algorithm \ref{algo:reduction}), that is ${NC_{t-9},\dots, NC_t}$. 
Here, $NC^r(i,j) = 0.1\sum_{k=0}^{9} NC_{t-9+k}(i,j)$, where $i,j \in [1, |\mathcal{E}|]$. In the original ORNCIE, only $NC_t$ is considered to detect the training process rather than the $NC^r$ we used. This second enhancement is because $NC^r$ can more precisely capture the correlations among objectives for the current model training.

In the loop, the algorithm identifies the most essential (here defined as conflicting) objective, denoted as $\mathcal{E}_J$. Subsequently, the objectives, referred to as $Del$, that are positively correlated with $\mathcal{E}_J$ are excluded (line \ref{line:cut} in Algorithm \ref{algo:reduction}).  
When determining $Del$, the original approach uses the ``classifying objectives'' strategy (shown in Algorithm 2 of the paper~\cite{wang2016objective}). Instead, we use a static hyperparameter $\tau$ to identify $Del$, which is the third enhancement strategy. The dynamic determination $Del$ using the ``classifying objectives'' strategy may wrongly remove some essential objectives and further degrade the representation ability of $\mathcal{E}'$.
Following this iterative process, a representative objective subset $\mathcal{E}'$ is determined. The appropriate setting of the unique hyperparameter $\tau$ will be discussed in Section \ref{sec:fairness_aware_sensitivity}.

\section{Experimental Studies}\label{sec:exp}
In this section, Section \ref{sec:overview} introduces the aims of our experimental studies. Then, Section \ref{sec:setting} presents the experimental settings.  Four experiments are presented and discussed in Section \ref{sec:framework_Effectiveness} to Section \ref{sec:Parameter_sensitivity}, respectively.

\subsection{Overview of Experimental Studies} \label{sec:overview}
Four experiments are conducted to achieve a comprehensive analysis of our proposed fairness-aware multiobjective evolutionary learning framework and its instantiation algorithm. Experimental setting is detailed in Section \ref{sec:setting}. First, we verify the effectiveness of our fairness-aware framework in Section \ref{sec:framework_Effectiveness} by comparing it with two state-of-the-art methods in multi-objective optimisation for fair machine learning: one that optimises the entire set of objectives~\cite{QingquanFair2021} and another that uses a static representative subset~\cite{Qingquan2022Ensemble}. Next, in Section \ref{sec:frequent_Effectiveness}, to further investigate the capabilities of our framework, we analyse whether the frequently selected objectives obtained by our algorithm for each dataset are more suitable as a new representative subset. Furthermore, in Section \ref{sec:fairness_aware_sensitivity}, ablation studies to assess the effectiveness of our fairness-aware strategy are conducted. 
Finally, we will analyse the sensitivity of the unique parameter $\tau$ of our method, as described in Algorithm \ref{algo:framework} and Algorithm \ref{algo:reduction}, and suggest a value for $\tau$.

\subsection{Experimental Setting}\label{sec:setting}

\subsubsection{Compared Methods}
A total of 26 objectives, including accuracy $CE$ and 25 fairness measures ($f_1$--$f_{25}$, described in Table \ref{tab:Fairness25}), are considered for all methods.
Our proposed method, denoted as $FaMOEL$, is described in Algorithm \ref{algo:framework} and Algorithm \ref{algo:reduction}. We compare $FaMOEL$ with two state-of-the-art algorithms, namely $MOEL$~\cite{QingquanFair2021} and $MOEL_{Rep}$~\cite{Qingquan2022Ensemble}, respectively. $MOEL$ directly optimises 26 objectives through the multiobjective evolutionary learning framework. $MOEL_{Rep}$ focuses on optimising a static representative subset. The work of~\cite{fairenough2023} clusters $f_1$--$f_{25}$ into six groups. The static representative subset can be identified by selecting a measure from each group. Our study selects $f_4$, $f_7$, $f_{10}$, $f_{16}$, $f_{17}$ and $f_{25}$ as the representative fairness measure subset. Thus, $MOEL_{Rep}$ optimises $CE$ and these six fairness objectives.

\subsubsection{Datasets}
12 well-known benchmark datasets widely used in the literature of  fairness~\cite{pessach2020algorithmic,fairenough2023} are considered,
namely
Heart health~\cite{chicco2020machine}, 
Titanic~\cite{titanic},
German~\cite{kamiran2009classifying}, 
Student performance~\cite{hussain2018classification},
COMPAS~\cite{compasdataset}, 
Bank~\cite{zafar2017fairness2},
Adult~\cite{adultdataset},
Drug~\cite{fehrman2017five},
Patient~\cite{Sadikin2020},
LSAT~\cite{sander2004systemic},  
Default~\cite{YEH20092473} and
Dutch~\cite{kamiran2012data}.
Table \ref{tab:dataset} summarises these datasets used.
Note that the analysis of the static representative objective subset for $MOEL_{Rep}$ is based on the first seven datasets~\cite{fairenough2023}, while the remaining five datasets are not included. This can help to further verify the effectiveness of our framework on the new benchmark datasets.
The pre-processing steps for Heart health, Titanic, German, Student performance, COMPAS, Bank, and Adult datasets follow the same procedure as described in~\cite{fairenough2023}. 
Each dataset is randomly divided into three partitions: a training set, a validation set, and a test set, with a split ratio of 6:2:2. All sensitive features listed in Table \ref{tab:dataset} are taken into consideration in the calculation of these objectives.

\subsubsection{Parameter Settings}
All the methods, including $FaMOEL$, $MOEL_{Rep}$ and $MOEL$, use the same settings for a fair comparison, introduced as follows. Each individual in the population is designed with a fully connected architecture, consisting of one hidden layer~\cite{Qingquan2022Ensemble,QingquanFair2021}. The values of the learning rate, mutation strength and the number of hidden nodes are determined using grid search and are summarised in Table \ref{tab:parameters}. The multiobjective optimiser $\pi$ is Two\_Arch2~\cite{wang2015twoarch2}. The size of the convergence and diversity archives are all set as 100. We set the termination condition to a maximum of $100$ generations. In our fairness-aware method $FaMOEL$, the selection threshold $\tau$ is set to 0.22. Five-fold cross-validation is used in our experiments, where 10 trials are independently run for each fold. Thus, 50 trials in total are performed for each benchmark dataset.

\begin{table}[htbp]
  \centering
  \caption{Parameter settings of algorithms for each dataset}
    \begin{tabular}{cccc}
    \toprule
    Dataset & Learning Rate & Mutation Strength & \#Nodes \\
    \midrule
    Heart health & 0.0001 & 0.0001 & 16 \\
    Titanic & 0.001 & 0.0001 & 8 \\
    German & 0.0001 & 0.05  & 64 \\
    Student performance & 0.001 & 0.0001 & 64 \\
    COMPAS & 0.001 & 0.05  & 64 \\
    Bank  & 0.001 & 0.005 & 64 \\
    Adult & 0.001 & 0.05  & 64 \\
    Drug consumption & 0.001 & 0.0001 & 64 \\
    Patient treatment & 0.0001 & 0.0001 & 64 \\
    LSAT  & 0.001 & 0.005 & 64 \\
    Default & 0.001 & 0.01  & 64 \\
    Dutch & 0.001 & 0.01  & 64 \\
    \bottomrule
    \end{tabular}%
  \label{tab:parameters}%
\end{table}%

\subsubsection{Performance Measures}

The quality of a model set can be assessed from four aspects, including convergence, spread, uniformity, and cardinality~\cite{li2019quality}. For a more comprehensive analysis, we adopt widely used generational distance (GD)~\cite{van1998evolutionary} for convergence, pure diversity (PD)~\cite{wang2016diversity} for spread, and spacing (SP)~\cite{schott1995fault} for uniformity, as suggested in~\cite{li2019quality,ZHANG2023101405}, and summarise them in Table V. PD and SP can collectively depict the diversity of a solution set. A solution set with diverse performance not only provides decision-makers with a better understanding of the task at hand but also offers flexibility, allowing them to choose the most suitable ones according to varying requirements. Moreover, hypervolume (HV)~\cite{zitzler1998multiobjective}, also known as the only indicator with Pareto compliance~\cite{ishibuchi2019comparison}, is applied to measure the overall performance. All the objective values are normalised before computing HV values The reference point $(1.2,1.2,\dots,1.2)$ is used for HV. A larger PD or HV value indicates better performance with respect to its corresponding property, while a smaller GD or SP value indicates better performance.

\begin{table}[htbp]
  \caption{Quality indicators for evaluating a solution set }
  \centering
  \begin{adjustbox} {max width=1\linewidth}
  \begin{tabular}{ccccc}
  \toprule
   Quality indicator & Convergence & Spread & Uniformity & Cardinality\\
   \midrule
   Generational distance (GD) & $\checkmark$ & & &\\
   Pure diversity (PD) & & $\checkmark$ & &\\
   Spacing (SP) & & &$\checkmark$ &\\
   Hypervolume (HV) & $\checkmark$ & $\checkmark$ & $\checkmark$ & $\checkmark$\\
  \bottomrule
  \end{tabular}
  \end{adjustbox}
  \label{tab:indicators}
\end{table}

\subsection{Effectiveness of Our Framework}\label{sec:framework_Effectiveness}
\begin{figure}[htbp]
  \begin{center}  \includegraphics[width=0.5\textwidth]{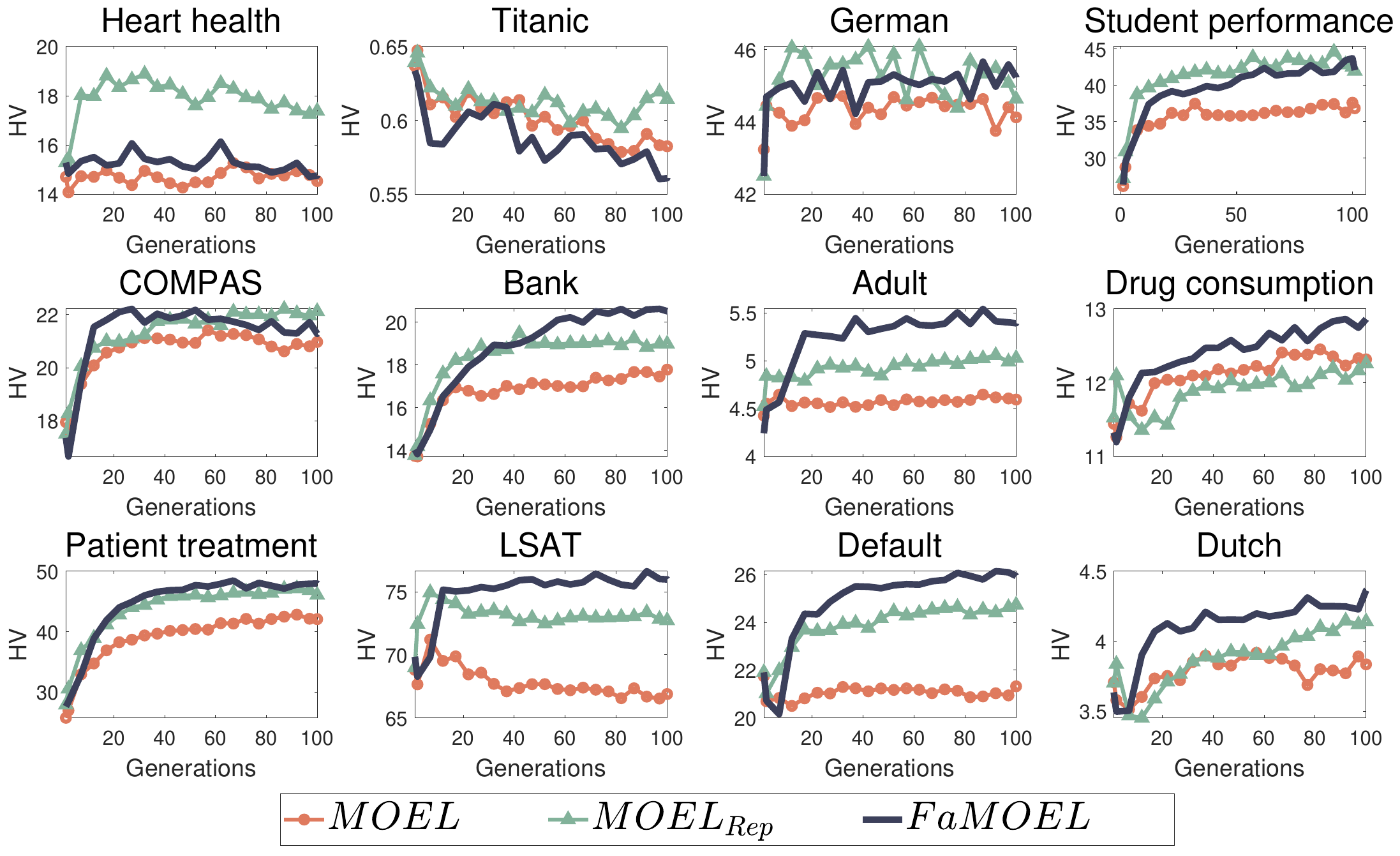}
  \end{center}
  \caption{HV curves along with generations averaged over 50 trials considering accuracy and $f_1$--$f_{25}$
  } \label{fig:HV_curve}
\end{figure}

To verify the effectiveness of our fairness-aware MOEL framework, three perspectives are considered, (i) convergence curves of HV values of $MOEL$, $MOEL_{Rep}$ and our proposed $FaMOEL$, (ii) Quality of population of the final generation in terms of GD, PD, SP and HV, (iii) Visualisation of the fairness-aware process of our proposed $FaMOEL$,

Fig. \ref{fig:HV_curve} illustrates the convergence curves of HV values considering $MOEL$, $MOEL_{Rep}$ and our proposed $FaMOEL$ on the test data. In the calculation of HV, the pseudo Pareto front is the non-dominated model set with respect to 26 objectives by considering all the models obtained by the three compared algorithms from all the generations across 50 trials, as suggested in~\cite{Qingquan2022Ensemble}. Fig. \ref{fig:HV_curve} reveals that our proposed method $FaMOEL$ (black) is better than $MOEL$ (orange) and $MOEL_{Rep}$ (green) on 7 out of 12 datasets, including Bank, Adult, Drug consumption, Patient treatment, LSAT, Default and Dutch. It means that $FaMOEL$ can achieve better overall performance in terms of $CE$ and $f_1$--$f_{25}$ and outperforms the state-of-the-art algorithms on these seven datasets, resulting from the fairness-aware strategy.
The poor performance of $FaMOEL$ on the remaining five datasets may be partially attributed to the small size~\cite{wang2020generalizing} of the datasets. A closer examination is expected in the future.

The curves of $MOEL_{Rep}$ exhibit an increasing trend with the generations on 7 out of 12 datasets including Student performance, COMPAS, Bank, Adult, Patient treatment, Default and Dutch. Additionally, $MOEL_{Rep}$ performs better than $MOEL$ on these datasets. The observation suggests that the subset can represent the entire objectives to some extent. However, this subset used in $MOEL_{Rep}$ may not serve as an optimal representative subset to guide the model training when compared to our proposed $FaMOEL$. In contrast, $FaMOEL$ leverages fairness awareness based on the current evolution process to select a more suitable subset representing the 26 objectives. Therefore, the adaptively determined subset instead of a pre-defined static one is more suitably used to guide model training.

Additionally, the distribution properties of the final solution set, including convergence, spread and uniformity, are used in further analysis, as summarised in Table \ref{tab:four_indicator1}.
In terms of overall performance measured by HV, $FaMOEL$ outperforms both $MOEL_{Rep}$ and $MOEL$. 
The pairwise win/tie/loss counts of $MOEL_{Rep}$ and $MOEL$ against  $FaMOEL$ are 2/4/6 and 0/3/9, respectively.

\begin{table*}[htbp]
  \centering
  \caption{GD, PD, SP and HV values of final model set averaged over 50 trials. ``+/$\approx$/-'' indicates that the average indicator value of the corresponding algorithm (specified by column header) is statistically better/similar/worse than the one of $FaMOEL$ according to the Friedman test with a 0.05 significance level. The best averaged values are highlighted in grey.}
  \setlength{\tabcolsep}{3pt}
\begin{tabular}{cccc|ccc}
    \toprule 
     \multirow{2}{*}{Dataset} & \multicolumn{3}{|c|}{\textbf{GD}} & \multicolumn{3}{c}{\textbf{PD}} \\
       & \multicolumn{1}{|c}{$MOEL$} & \multicolumn{1}{c}{$MOEL_{Rep}$} & \multicolumn{1}{c|}{$FaMOEL$} & \multicolumn{1}{c}{$MOEL$} & \multicolumn{1}{c}{$MOEL_{Rep}$} & \multicolumn{1}{c}{$FaMOEL$} \\
    \midrule
    \multicolumn{1}{c|}{Heart health} & \multicolumn{1}{c}{2.96e-04(8.9e-05)$\approx$} & \multicolumn{1}{c}{\cellcolor[rgb]{ .851  .851  .851}2.56e-04(8.6e-05)+} & \multicolumn{1}{l|}{2.88e-04(8.3e-05)} & \multicolumn{1}{l}{\cellcolor[rgb]{ .851  .851  .851}2.31e+14(4.7e+13)+} & \multicolumn{1}{l}{1.45e+14(5.0e+13)-} & \multicolumn{1}{l}{2.08e+14(4.5e+13)} \\
    \multicolumn{1}{c|}{Titanic} & \multicolumn{1}{l}{1.36e-04(8.7e-05)-} & \multicolumn{1}{l}{1.28e-04(8.3e-05)$\approx$} & \multicolumn{1}{c|}{\cellcolor[rgb]{ .851  .851  .851}1.25e-04(8.8e-05)} & \multicolumn{1}{l}{\cellcolor[rgb]{ .851  .851  .851}4.72e+14(3.8e+13)+} & \multicolumn{1}{l}{4.31e+14(3.3e+13)$\approx$} & \multicolumn{1}{l}{4.40e+14(4.0e+13)} \\
    \multicolumn{1}{c|}{German} & \multicolumn{1}{l}{5.66e-04(2.2e-04)-} & \multicolumn{1}{l}{\cellcolor[rgb]{ .851  .851  .851}4.85e-04(1.6e-04)$\approx$} & \multicolumn{1}{c|}{5.24e-04(2.1e-04)} & \multicolumn{1}{l}{\cellcolor[rgb]{ .851  .851  .851}4.11e+14(1.0e+14)+} & \multicolumn{1}{l}{3.71e+14(1.2e+14)-} & \multicolumn{1}{l}{3.92e+14(1.1e+14)} \\
    \multicolumn{1}{c|}{Student performance} & \multicolumn{1}{l}{6.11e-04(1.7e-04)-} & \multicolumn{1}{l}{\cellcolor[rgb]{ .851  .851  .851}4.66e-04(1.2e-04)+} & \multicolumn{1}{c|}{5.04e-04(1.5e-04)} & \multicolumn{1}{l}{\cellcolor[rgb]{ .851  .851  .851}4.51e+14(7.7e+13)+} & \multicolumn{1}{l}{3.77e+14(7.7e+13)-} & \multicolumn{1}{l}{4.06e+14(9.3e+13)} \\
    \multicolumn{1}{c|}{COMPAS} & \multicolumn{1}{l}{2.09e-04(6.7e-05)-} & \multicolumn{1}{l}{\cellcolor[rgb]{ .851  .851  .851}1.87e-04(5.7e-05)+} & \multicolumn{1}{c|}{2.00e-04(6.6e-05)} & \multicolumn{1}{l}{\cellcolor[rgb]{ .851  .851  .851}4.24e+14(5.0e+13)+} & \multicolumn{1}{l}{3.99e+14(4.7e+13)$\approx$} & \multicolumn{1}{l}{4.07e+14(4.6e+13)} \\
    \multicolumn{1}{c|}{Bank} & \multicolumn{1}{l}{9.41e-05(8.3e-05)-} & \multicolumn{1}{l}{\cellcolor[rgb]{ .851  .851  .851}7.94e-05(6.7e-05)$\approx$} & \multicolumn{1}{c|}{8.02e-05(7.8e-05)} & \multicolumn{1}{l}{\cellcolor[rgb]{ .851  .851  .851}4.47e+14(6.2e+13)+} & \multicolumn{1}{l}{4.09e+14(5.4e+13)$\approx$} & \multicolumn{1}{l}{4.15e+14(6.4e+13)} \\
    \multicolumn{1}{c|}{Adult} & \multicolumn{1}{l}{5.39e-05(3.3e-05)-} & \multicolumn{1}{l}{\cellcolor[rgb]{ .851  .851  .851}4.51e-05(3.3e-05)+} & \multicolumn{1}{c|}{5.03e-05(3.1e-05)} & \multicolumn{1}{l}{\cellcolor[rgb]{ .851  .851  .851}5.78e+14(1.9e+13)+} & \multicolumn{1}{l}{5.04e+14(1.8e+13)-} & \multicolumn{1}{l}{5.34e+14(1.9e+13)} \\
    \multicolumn{1}{c|}{Drug consumption} & \multicolumn{1}{l}{1.74e-04(9.9e-05)-} & \multicolumn{1}{l}{1.69e-04(1.1e-04)$\approx$} & \multicolumn{1}{c|}{\cellcolor[rgb]{ .851  .851  .851}1.54e-04(8.5e-05)} & \multicolumn{1}{l}{\cellcolor[rgb]{ .851  .851  .851}4.08e+14(4.4e+13)+} & \multicolumn{1}{l}{3.90e+14(5.0e+13)+} & \multicolumn{1}{l}{3.76e+14(3.8e+13)} \\
    \multicolumn{1}{c|}{Patient treatment} & \multicolumn{1}{l}{4.24e-04(2.0e-04)-} & \multicolumn{1}{l}{\cellcolor[rgb]{ .851  .851  .851}2.54e-04(8.1e-05)$\approx$} & \multicolumn{1}{c|}{2.77e-04(1.2e-04)} & \multicolumn{1}{l}{\cellcolor[rgb]{ .851  .851  .851}4.60e+14(1.3e+14)+} & \multicolumn{1}{l}{3.51e+14(9.3e+13)$\approx$} & \multicolumn{1}{l}{3.51e+14(1.1e+14)} \\
    \multicolumn{1}{c|}{LSAT} & \multicolumn{1}{l}{2.23e-04(8.2e-05)-} & \multicolumn{1}{l}{\cellcolor[rgb]{ .851  .851  .851}1.49e-04(5.2e-05)+} & \multicolumn{1}{c|}{1.80e-04(7.2e-05)} & \multicolumn{1}{l}{\cellcolor[rgb]{ .851  .851  .851}2.98e+14(6.0e+13)+} & \multicolumn{1}{l}{2.34e+14(6.4e+13)$\approx$} & \multicolumn{1}{l}{2.36e+14(6.7e+13)} \\
    \multicolumn{1}{c|}{Default} & \multicolumn{1}{l}{6.93e-05(3.1e-05)-} & \multicolumn{1}{l}{\cellcolor[rgb]{ .851  .851  .851}5.30e-05(2.6e-05)+} & \multicolumn{1}{c|}{5.78e-05(2.9e-05)} & \multicolumn{1}{l}{\cellcolor[rgb]{ .851  .851  .851}3.46e+14(2.0e+13)+} & \multicolumn{1}{l}{3.07e+14(1.8e+13)$\approx$} & \multicolumn{1}{l}{3.04e+14(1.7e+13)} \\
    \multicolumn{1}{c|}{Dutch} & \multicolumn{1}{l}{5.22e-05(1.3e-05)-} & \multicolumn{1}{l}{\cellcolor[rgb]{ .851  .851  .851}2.76e-05(1.2e-05)+} & \multicolumn{1}{c|}{3.85e-05(1.4e-05)} & \multicolumn{1}{l}{\cellcolor[rgb]{ .851  .851  .851}6.14e+14(2.1e+13)+} & \multicolumn{1}{l}{5.55e+14(2.0e+13)-} & \multicolumn{1}{l}{5.76e+14(2.3e+13)} \\
    \midrule
      \multicolumn{1}{c|}{+/$\approx$/-}      & 0/1/11      &   7/4/0    &   -    &   12/0/0    &   1/6/5    &  -\\
    \bottomrule
    
     \\
     
     \toprule
          \multirow{2}{*}{Dataset} & \multicolumn{3}{|c|}{\textbf{SP}} & \multicolumn{3}{c}{\textbf{HV}} \\
           & \multicolumn{1}{|c}{$MOEL$} & \multicolumn{1}{c}{$MOEL_{Rep}$} & \multicolumn{1}{c|}{$FaMOEL$} & \multicolumn{1}{c}{$MOEL$} & \multicolumn{1}{c}{$MOEL_{Rep}$} & \multicolumn{1}{c}{$FaMOEL$} \\
    \midrule
    \multicolumn{1}{c|}{Heart health} & 0.661(9.1e-02)$\approx$ & 0.678(0.15)- & \cellcolor[rgb]{ .851  .851  .851 }0.660(9.8e-02) & 14.7(8.5)$\approx$ & \cellcolor[rgb]{ .851  .851  .851 }17.4(9.4)+ & 15.1(8.3) \\
    \multicolumn{1}{c|}{Titanic} & \cellcolor[rgb]{ .851  .851  .851 }0.590(5.2e-02)$\approx$ & 0.600(6.5e-02)$\approx$ & 0.597(5.9e-02) & 0.586(0.35)$\approx$ & \cellcolor[rgb]{ .851  .851  .851 }0.621(0.36)+ & 0.578(0.37) \\
    \multicolumn{1}{c|}{German}  & 0.718(0.13)$\approx$ & \cellcolor[rgb]{ .851  .851  .851 }0.700(0.12)$\approx$ & 0.719(0.14) & 43.6(30)- & \cellcolor[rgb]{ .851  .851  .851 }45.4(31)$\approx$ & 44.2(30) \\
    \multicolumn{1}{c|}{{Student} performance}  & 0.706(0.13)$\approx$ & \cellcolor[rgb]{ .851  .851  .851 }0.679(0.12)$\approx$ & 0.702(0.13) & 36.8(21)- & 41.9(22)$\approx$ & \cellcolor[rgb]{ .851  .851  .851 }42.0(23) \\
    \multicolumn{1}{c|}{COMPAS}  & 0.529(7.3e-02)$\approx$ & 0.532(7.2e-02)$\approx$ & \cellcolor[rgb]{ .851  .851  .851 }0.526(8.0e-02) & 20.7(8.6)$\approx$ & \cellcolor[rgb]{ .851  .851  .851 }22.2(9.9)$\approx$ & 21.4(8.6) \\
    \multicolumn{1}{c|}{Bank}  & 0.528(6.6e-02)$\approx$ & \cellcolor[rgb]{ .851  .851  .851 }0.481(7.8e-02)+ & 0.528(7.4e-02) & 17.4(5.0)- & 19.0(4.9)- & \cellcolor[rgb]{ .851  .851  .851 }20.5(5.8) \\
    \multicolumn{1}{c|}{Adult}  & 0.492(5.3e-02)$\approx$ & \cellcolor[rgb]{ .851  .851  .851 }0.458(5.3e-02)+ & 0.500(5.2e-02) & 4.58(0.79)- & 4.99(0.70)- & \cellcolor[rgb]{ .851  .851  .851 }5.48(1.1) \\
    \multicolumn{1}{c|}{Drug consumption}  & 0.485(5.7e-02)$\approx$ & \cellcolor[rgb]{ .851  .851  .851 }0.455(5.8e-02)+ & 0.486(6.1e-02) & 12.3(5.7)- & 12.2(5.7)- & \cellcolor[rgb]{ .851  .851  .851 }12.9(5.7) \\
    \multicolumn{1}{c|}{Patient treatment}  & 0.709(0.13)$\approx$ & \cellcolor[rgb]{ .851  .851  .851 }0.639(0.15)$\approx$ & 0.663(0.17) & 42.3(14)- & 45.9(15)$\approx$ & \cellcolor[rgb]{ .851  .851  .851 }47.3(15) \\
    \multicolumn{1}{c|}{LSAT}  & 0.439(5.1e-02)$\approx$ & \cellcolor[rgb]{ .851  .851  .851 }0.383(0.11)+ & 0.456(7.7e-02) & 67.0(9.59)- & 72.9(9.3)- & \cellcolor[rgb]{ .851  .851  .851 }75.6(11) \\
    \multicolumn{1}{c|}{Default}  & 0.408(5.3e-02)$\approx$ & \cellcolor[rgb]{ .851  .851  .851 }0.369(5.5e-02)+ & 0.412(5.1e-02) & 21.2(2.34)- & 24.6(2.72)- & \cellcolor[rgb]{ .851  .851  .851 }25.7(2.5) \\
    \multicolumn{1}{c|}{Dutch}  & 0.523(5.8e-02)- & 0.483(5.8e-02)$\approx$ & \cellcolor[rgb]{ .851  .851  .851 }0.463(5.3e-02) & 3.75(0.51)- & 4.17(0.46)- & \cellcolor[rgb]{ .851  .851  .851 }4.37(0.66) \\
    \midrule
      \multicolumn{1}{c|}{+/$\approx$/-}    &   0/11/1    & 5/6/1     &    -   &  0/3/9     &  2/4/6     &  - \\
    \bottomrule
    \end{tabular}%
  \label{tab:four_indicator1}%
\end{table*}%

Since $MOEL$ optimises all the 26 objectives, the effectiveness of $MOEL$ may degrade, leading to worse GD but better PD in comparison with $FaMOEL$. 
This can be attributed to the fact that a solution set that is far away from the Pareto front may exhibit worse convergence but better spread. 
However, for the model set obtained by $MOEL$, a significant loss in convergence (GD) leads to worse overall performance (HV).
On the contrary, $MOEL_{Rep}$ demonstrates better GD but worse PD, indicating that the model set obtained by $MOEL_{Rep}$ converges to a subregion of the Pareto front. That's because $MOEL_{Rep}$ only optimises the static subset and does not care about other considered objectives, which may make the models trap in the local regions considering all the objectives. What's more, on the seven datasets that are utilised to find a representative subset in $MOEL_{Rep}$, our proposed $FaMOEL$ performs no worse than $MOEL_{Rep}$ on 5 out of 7 datasets without any prior knowledge. Considering all the remaining datasets, $FaMOEL$ achieves better performance than $MOEL_{Rep}$ on 4 out of 5 datasets and no worse than $MOEL_{Rep}$ on the other dataset. This observation further verifies the effectiveness of our fairness-aware strategy using MOEL. More appropriate representative objectives can be constructed by $FaMOEL$ to guide the model training process.

In Titanic and Heart health datasets, $FaMOEL$ has worse HV performance compared to $MOEL_{Rep}$. In Titanic, although $FaMOEL$ and $MOEL_{Rep}$ have the same performance in terms of GD, PD and SP, $FaMOEL$ has a slightly worse HV performance. One possible reason is that HV may have a bias when measuring the overall performance~\cite{ishibuchi2019comparison} between the solution sets with very close convergence, spread and uniformity. For Heart health, $MOEL_{Rep}$ demonstrates better convergence and further achieves better overall performance, albeit with slightly worse PD and SP.

\begin{figure*}[htbp]
    \centering
    
    \includegraphics[width=.33\textwidth]{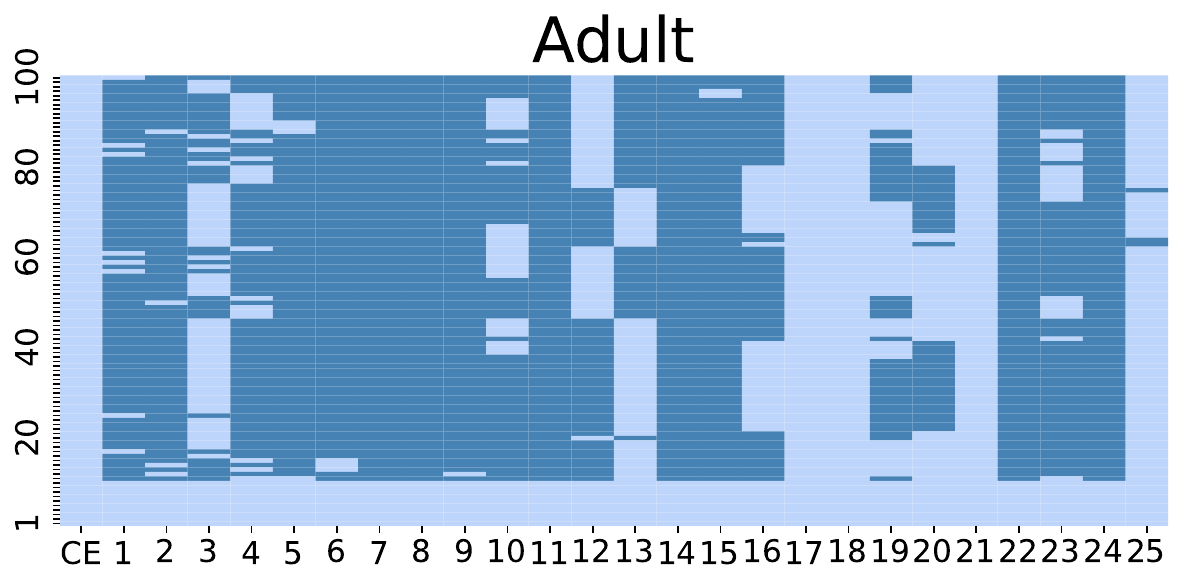}
    \includegraphics[width=.33\textwidth]{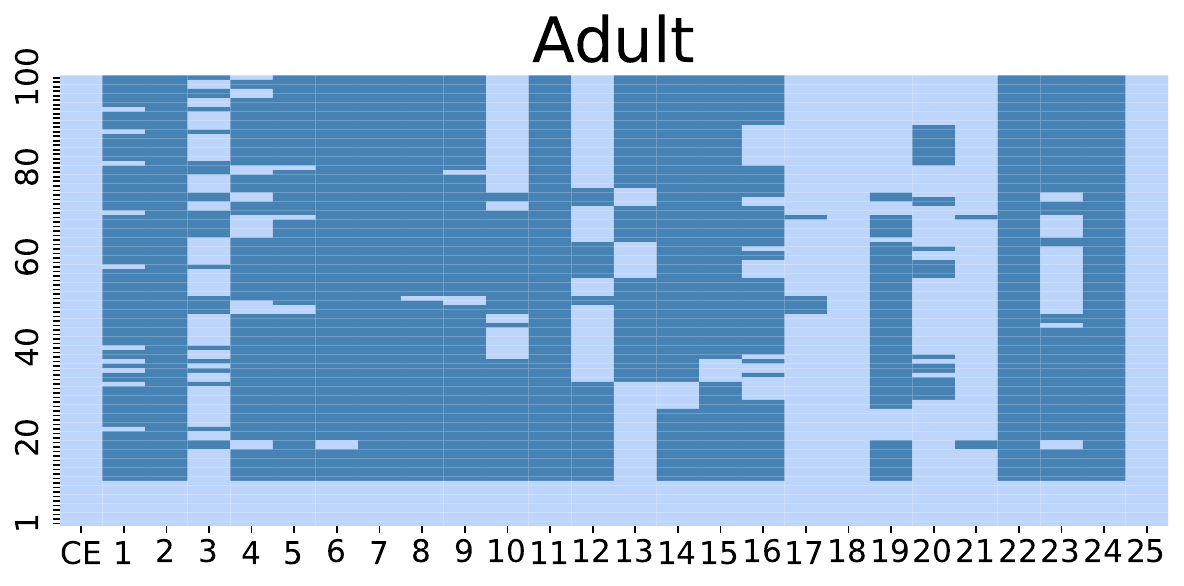}
    \includegraphics[width=.32\textwidth]{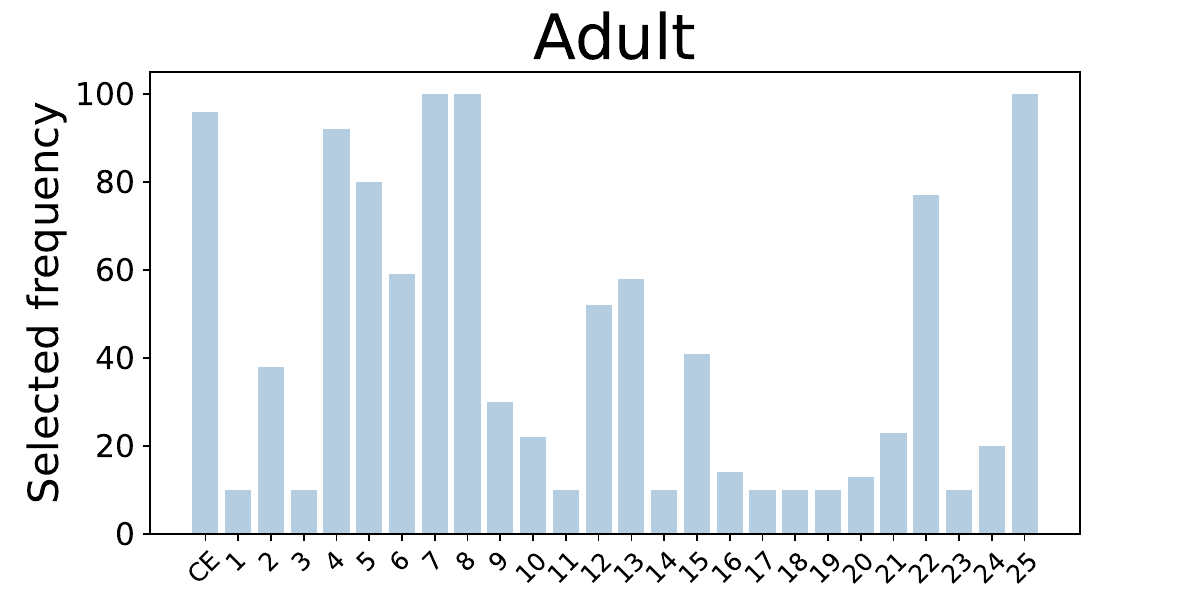}\\
    
    \includegraphics[width=.33\textwidth]{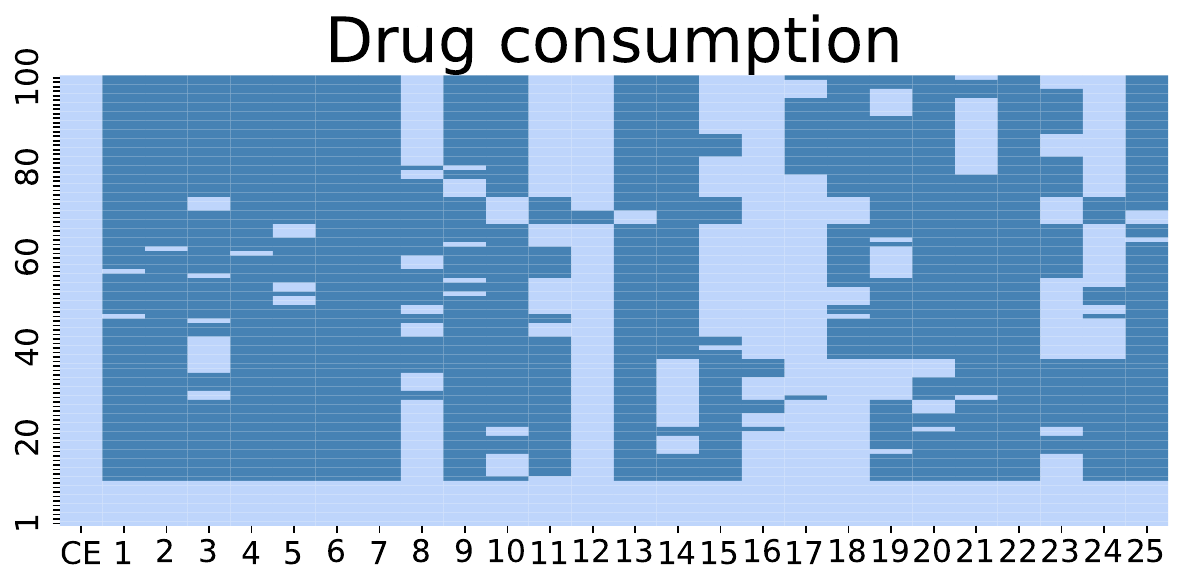}
    \includegraphics[width=.33\textwidth]{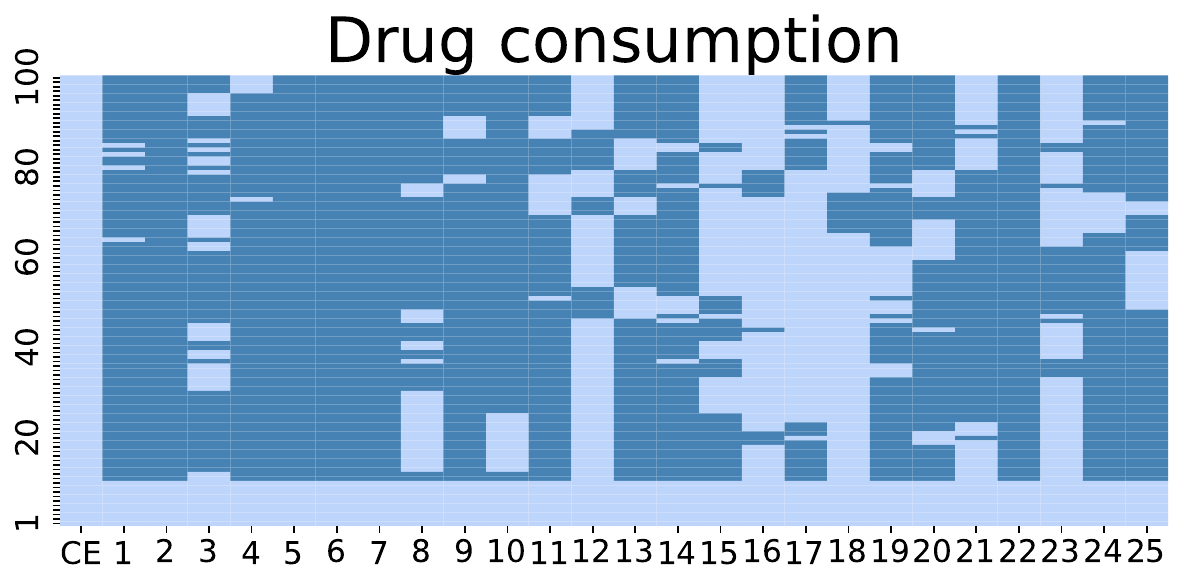}
    \includegraphics[width=.32\textwidth]{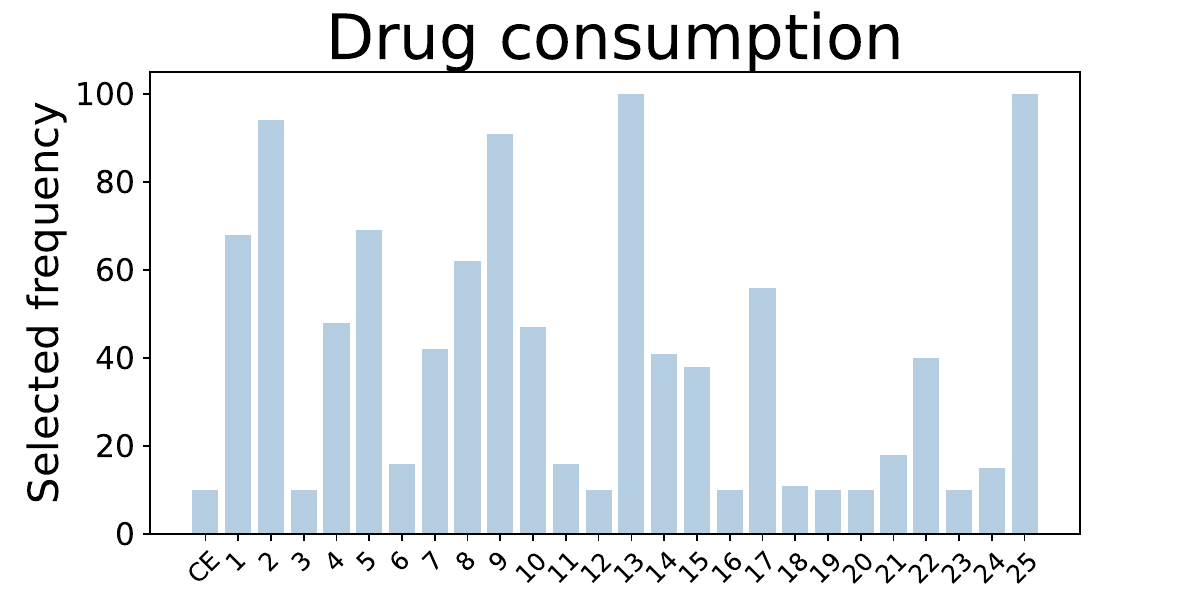}\\
    
    \includegraphics[width=.33\textwidth]{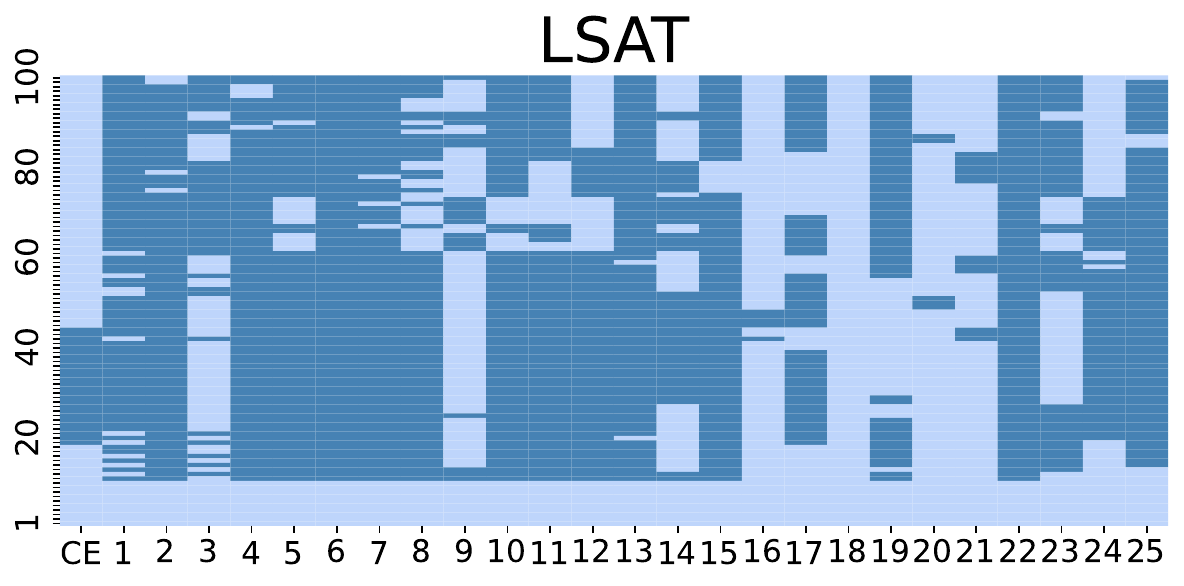}
    \includegraphics[width=.33\textwidth]{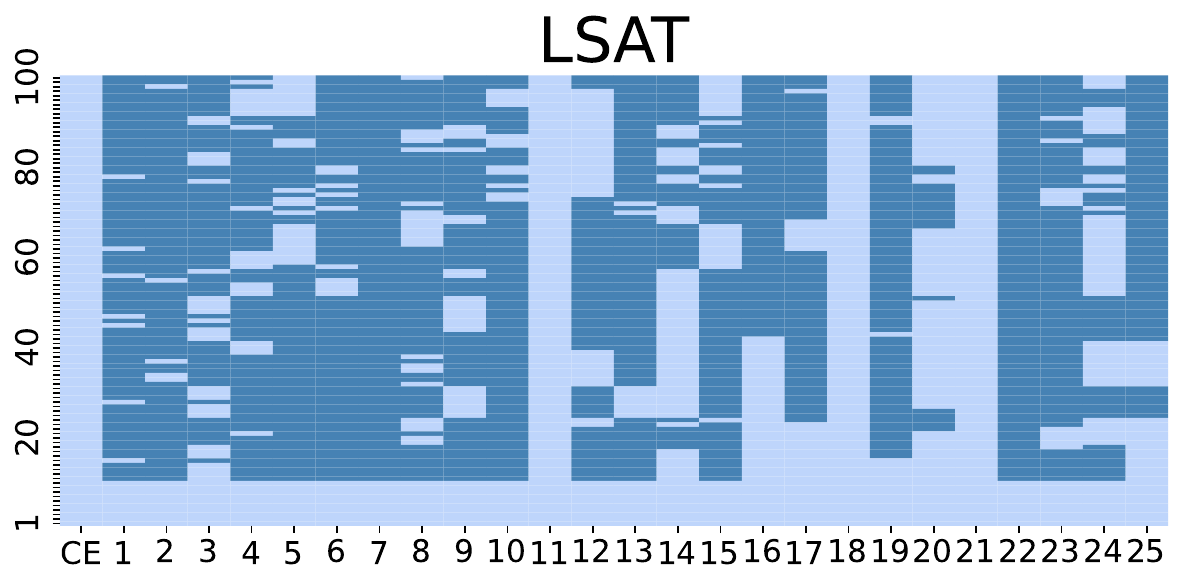}
    \includegraphics[width=.32\textwidth]{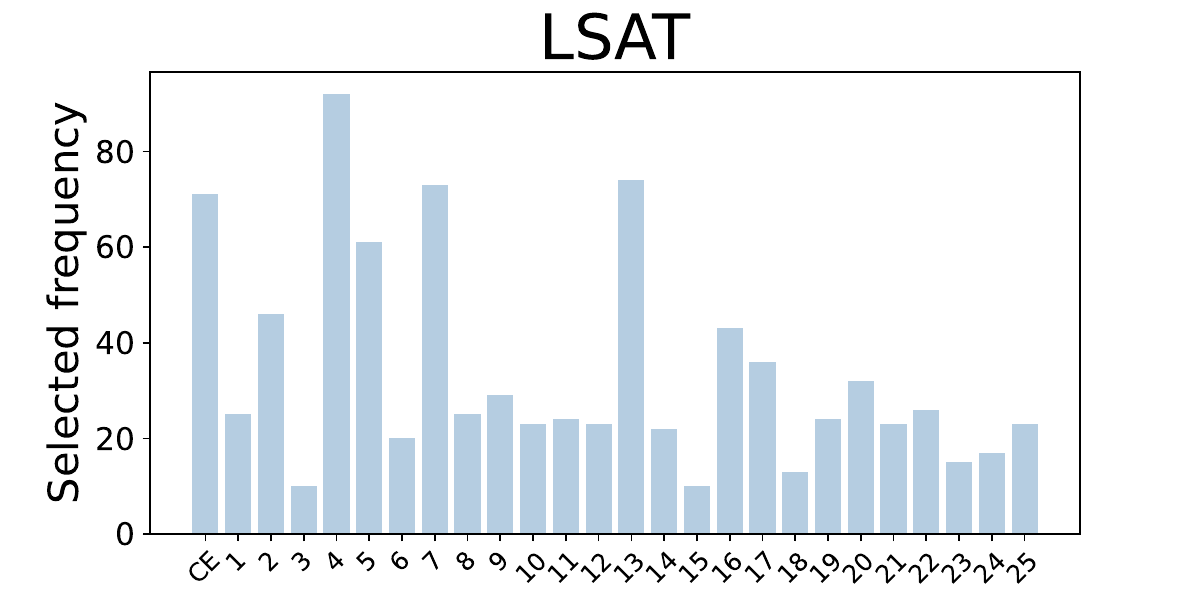}\\
    
    \caption{\label{fig:process}Visualisation of the fairness awareness process of $FaMOEL$. The first and second columns depict the evolution of the representative objective subset to be optimised at each generation. Each light-colored block represents the selection of an objective (corresponding to its respective column) for optimisation at the corresponding generation (corresponding to the row). The third column displays the average frequency of selecting each objective along with 100 generations over 50 trials.}
\end{figure*}

Then, we take a closer look at the fairness awareness process of our proposed $FaMOEL$. We present the results for Adult, Drug consumption, and LSAT datasets as examples, but the conclusions drawn from these datasets can be generalized to the remaining datasets as well. The visualisation of the fairness awareness process is depicted in Fig. \ref{fig:process}. The first and second columns in Fig. \ref{fig:process} illustrate the representative subset selected as optimisation objectives at each generation, where a light-colored block indicates that the objective (associated with its respective row) is selected for optimisation in the corresponding generation (determined by its column). Two arbitrary trials are plotted for demonstration purposes. Additionally, the averaged frequency of objective selection over 50 trials is presented in the last column of Fig. \ref{fig:process}, providing an overview of the selection patterns across the entire trial.

\begin{figure}[htbp]
  \begin{center}  \includegraphics[width=0.49\textwidth]{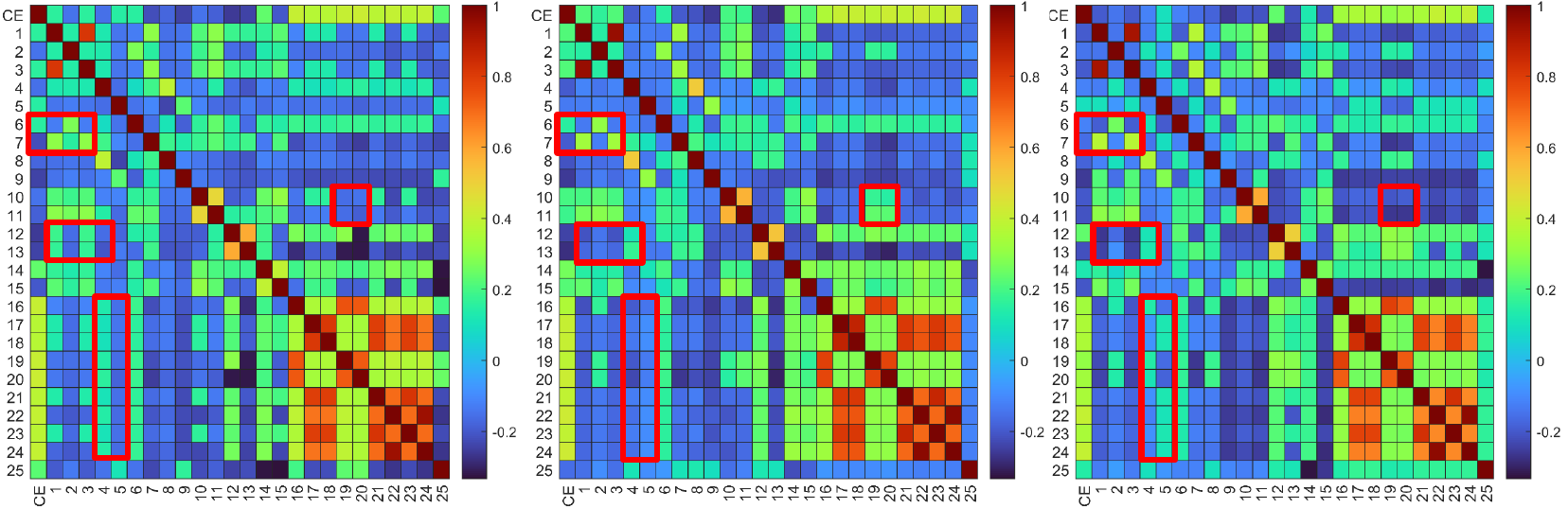}
  \end{center}
  \caption{Heatmap illustrating the correlation among accuracy and 25 fairness measures at generations 1, 50 and 100, respectively, in dealing with Drug consumption.
  } \label{fig:corr_heatmap_change}
\end{figure}

\begin{figure}[htbp]
  \begin{center} 
    \includegraphics[width=0.49\textwidth]{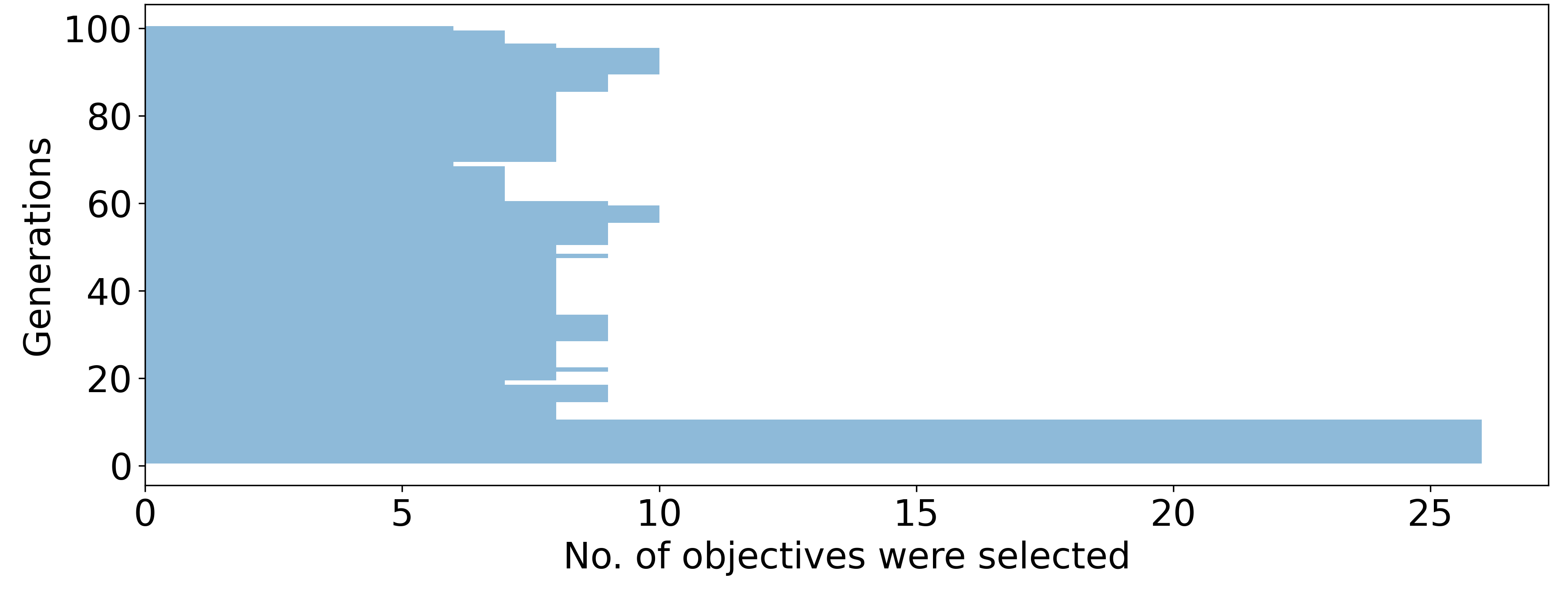}\\
    \includegraphics[width=0.49\textwidth]{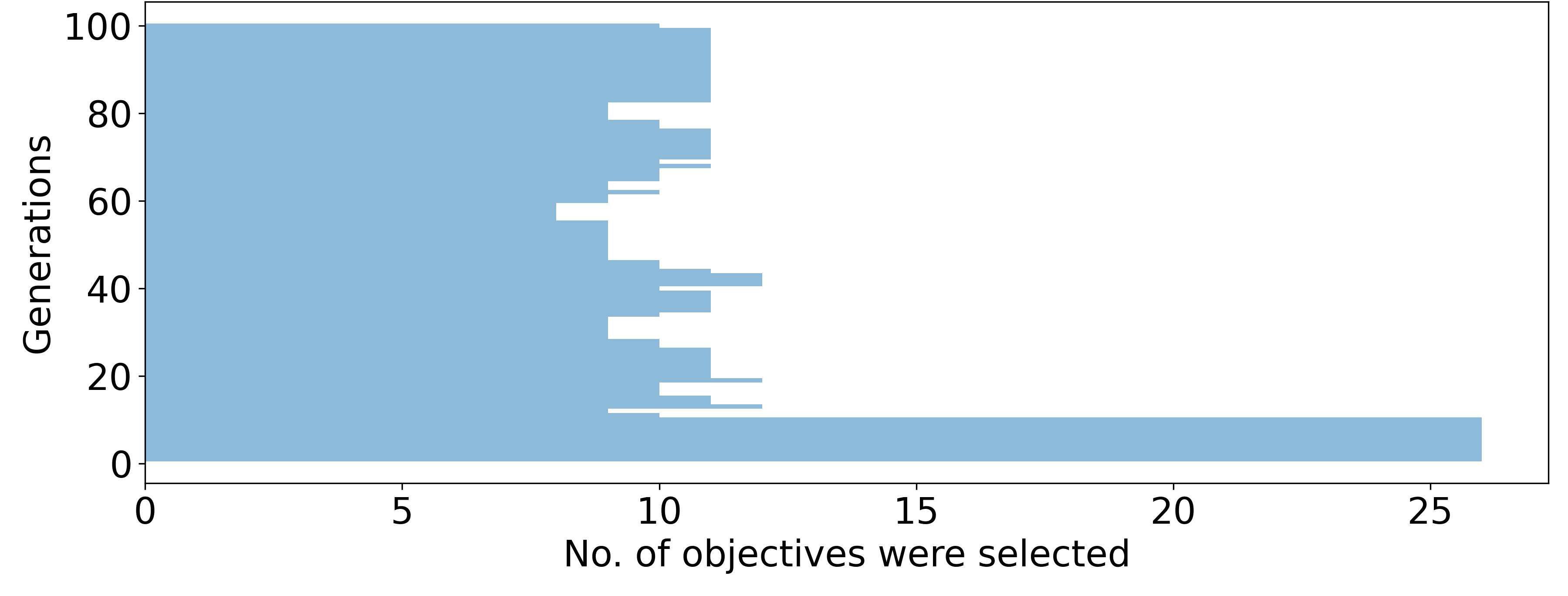}\\
  \end{center}
  \caption{Number of measures selected as objectives at each generation within two arbitrary trials on LSAT dataset.
    } \label{fig:number_opted_LSAT}
\end{figure}

Fig.~\ref{fig:corr_heatmap_change} illustrates the correlation among accuracy and 25 fairness measures at generations 1, 50 and 100 in a single trial when dealing with the dataset Drug consumption. A correlation value close to 1 indicates a stronger positive correlation, while a value close to -1 suggests a stronger negative correlation. The findings reveal dynamic changes in the correlations among measures, highlighted by the red boxes, throughout the training process. For example, the correlations between Fair4 and Fair16--Fair24 vary across generations: they are positively correlated at generation 1, but negatively correlated at generation 50. Furthermore, we illustrate the size of the selected representative subset (i.e., number of objectives) across generations on the LSAT dataset in Fig.~\ref{fig:number_opted_LSAT}, demonstrating that the number of objectives varies from generation to generation.

Fig. \ref{fig:process}, Fig.~\ref{fig:corr_heatmap_change} and  Fig.~\ref{fig:number_opted_LSAT} highlight the capability of our framework to identify a better representative subset along with the model training process. These processes have three aspects: (i) across different model training stages within the same trial, (ii) across different trials conducted on the same dataset, and (iii) across different datasets. 
The adaptively determined representative subset obtained by our framework according to current training stages is more suitable as optimisation objectives.
Based on the experimental results, we conclude for these three cases as follows.
In cases (i) and (ii), optimising a static subset may lead to local regions and potentially result in worse performance. This can be observed by the stabilised curves of $MOEL_{Rep}$ in the later stages of evolution. It becomes crucial to identify a more suitable representative subset that can help overcome these local regions. The dynamic awareness of representative objectives in our proposed framework enables the exploration of different subsets, ultimately leading to improved performance.
Regarding case (iii), as demonstrated by previous research~\cite{garg2020fairness}, the relationships among fairness objectives can vary depending on the dataset characteristics. Consequently, there is no universally applicable ``perfect" subset that can adequately and comprehensively represent all the objectives across all datasets. It further emphasises the need for our fairness awareness approach that can tailor the representative subset to the specific dataset being considered.

$FaMOEL$ using the fairness-aware strategy can adaptively select a properly representative subset of objectives according to the current process to guide the evolution of the population, which also does not rely on any prior knowledge.

\subsection{Comparison with Optimising Frequently Selected objectives}\label{sec:frequent_Effectiveness}
As depicted in the last column of Fig. \ref{fig:process}, the selection frequencies of different objectives among $CE$ and $f_1$--$f_{25}$ vary across different datasets. It's intriguing to analyse the comparison between $FaMOEL$ and the method optimising only the frequently selected, denoted as $MOEL_{Rep}^-$. In this study, we specifically compare the performance of $FaMOEL$ with that of $MOEL_{Rep}^-$, where $MOEL_{Rep}^-$ optimises different objectives while maintaining the same settings for the remaining factors as in $MOEL_{Rep}$. For each dataset, the objectives whose frequency is larger than 60 among 10 generations are selected as the optimised objective set in $MOEL_{Rep}^-$, which is summarised in Table \ref{tab:frequency}. Table \ref{tab:four_indicator2} presents the results of GD, PD, SP and HV obtained by $MOEL_{Rep}^-$ and $FaMOEL$, respectively. The results indicate that on 7 out of 12 datasets, $FaMOEL$ outperforms $MOEL_{Rep}^-$, while on 11 out of 12 datasets, $FaMOEL$ achieves no worse performance than $MOEL_{Rep}^-$. Specifically, $FaMOEL$ demonstrates better convergence and spread performance, as observed from the GD and PD measures.

\begin{table}[htbp]
  \centering
  \caption{Frequently Selected objectives of $FaMOEL$, which are used as optimised objectives of $MOEL_{Rep}^-$}
    \begin{tabular}{cc}
    \toprule
    Dataset & Objectives \\
    \midrule
    Heart health & $CE$, $f_9$, $f_{11}$, $f_{15}$, $f_{25}$ \\
    Titanic & $CE$, $f_8$, $f_{11}$, $f_{15}$, $f_{25}$ \\
    German & $f_4$, $f_5$, $f_8$, $f_{10}$, $f_{11}$, $f_{22}$, $f_{25}$ \\
    Student performance & $CE$, $f_9$, $f_{10}$, $f_{11}$, $f_{22}$, $f_{25}$ \\
    COMPAS & $CE$, $f_2$, $f_7$, $f_8$, $f_9$, $f_{10}$, $f_{25}$ \\
    Bank  & $CE$, $f_2$, $f_7$, $f_8$, $f_9$,$f_{10}$, $f_{11}$, $f_{15}$, $f_{25}$ \\
    Adult & $CE$, $f_4$, $f_5$, $f_8$, $f_9$, $f_{22}$, $f_{25}$ \\
    Drug consumption & $CE$, $f_2$, $f_4$, $f_8$, $f_{10}$, $f_{11}$, $f_{25}$ \\
    Patient treatment & $CE$, $f_4$, $f_5$, $f_8$, $f_{22}$, $f_{25}$ \\
    LSAT  & $f_2$, $f_4$, $f_5$, $f_8$, $f_9$, $f_{10}$, $f_{25}$ \\
    Default & $CE$, $f_1$, $f_4$, $f_5$, $f_6$, $f_7$, $f_8$, $f_9$, $f_{10}$, $f_{22}$, $f_{25}$ \\
    Dutch & $CE$, $f_2$, $f_6$, $f_8$, $f_9$, $f_{10}$, $f_{15}$, $f_{25}$ \\
    \bottomrule
    \end{tabular}%
  \label{tab:frequency}%
\end{table}%

\begin{figure}[htbp]
  \begin{center}  \includegraphics[width=.45\textwidth]{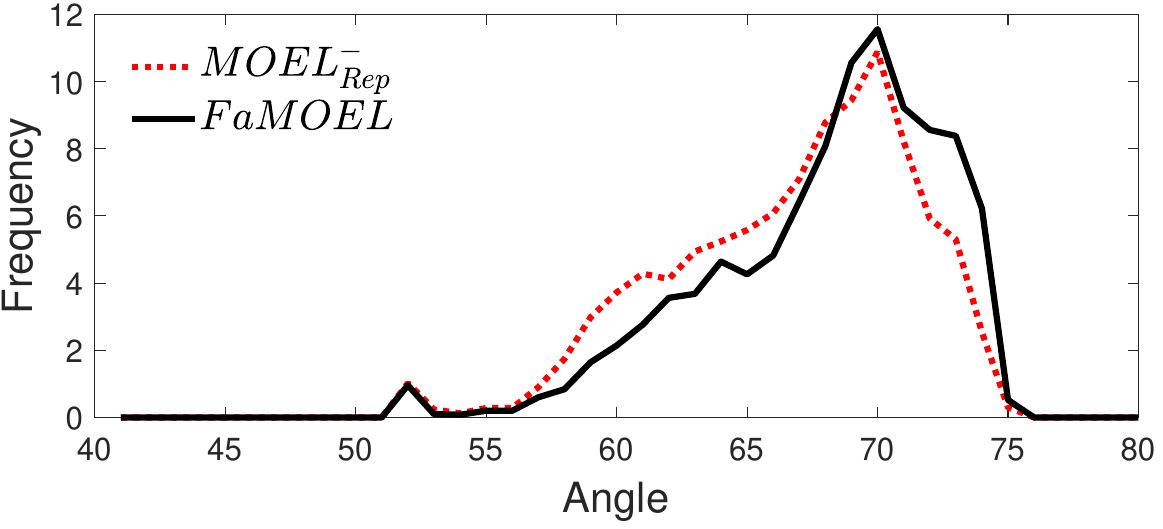}
  \end{center}
  \caption{Averaged frequency of minimal angles between each axis and each model of the final model set in the Dutch dataset over 50 trials
  } \label{fig:angle}
\end{figure}

In Dutch dataset, $FaMOEL$ performs worse than $MOEL_{Rep}^-$ in terms of HV. As Table \ref{tab:four_indicator2} shows, $FaMOEL$ performs better in GD and PD and the same in SP. We further measure the distribution property of the extremeness of the model set. For a model in each trial, the minimal angle, denoted as $a$, between the objective values and each axis is viewed as the extremeness of the model. A model with $a$ closer to $45^{\circ}$ indicates less serious extremeness and is a more centred distribution. Then, in Fig. \ref{fig:angle}, we plot the frequency histogram of angle $a$ of each model in the final set obtained by $MOEL_{Rep}^-$ and $FaMOEL$ over 50 trials, respectively. Fig. \ref{fig:angle} shows that the model sets obtained by $MOEL_{Rep}^-$ (black line) are more around the centre area, which contributes to better HV performance~\cite{ishibuchi2019comparison}.

In summary, the above observations validate the effectiveness of $FaMOEL$. The representative subset should be adaptively determined according to the model training stage.

\begin{table*}[htbp]
  \centering
  \caption{GD, PD, SP and HV values of final model set averaged over 50 trials. ``+/$\approx$/-'' indicates that the average indicator value of the corresponding algorithm (specified by column header) is statistically better/similar/worse than the one of $FaMOEL$ according to the Friedman test with a 0.05 significance level. The best averaged values are highlighted in grey.}
  \setlength{\tabcolsep}{1pt}
    \begin{tabular}{c|cc|cc|cc|cc}
    \toprule
       \multirow{2}{*}{Dataset}   & \multicolumn{2}{c|}{\textbf{GD}} & \multicolumn{2}{c|}{\textbf{PD}} & \multicolumn{2}{c|}{\textbf{SP}} & \multicolumn{2}{c}{\textbf{HV}} \\
    \multicolumn{1}{c|}{} & \multicolumn{1}{c}{$MOEL_{Rep}^-$} & \multicolumn{1}{c|}{$FaMOEL$} & \multicolumn{1}{c}{$MOEL_{Rep}^-$} & \multicolumn{1}{c|}{$FaMOEL$} & \multicolumn{1}{c}{$MOEL_{Rep}^-$} & \multicolumn{1}{c|}{$FaMOEL$} & \multicolumn{1}{c}{$MOEL_{Rep}^-$} & \multicolumn{1}{c}{$FaMOEL$} \\
    \midrule
   Heart health & \cellcolor[rgb]{ .851  .851  .851}2.66e-04(9.3e-05)+ & 2.88e-04(8.4e-05) & 1.42e+14(3.1e+13)- & \cellcolor[rgb]{ .851  .851  .851}2.08e+14(4.5e+13) & 0.710(9.6e-02)- & \cellcolor[rgb]{ .851  .851  .851}0.660(9.8e-02) & 10.0(7.7)- & \cellcolor[rgb]{ .851  .851  .851}15.1(8.3) \\
    Titanic & 1.39e-04(7.8e-05)- & \cellcolor[rgb]{ .851  .851  .851}1.25e-04(8.9e-05) & 4.04e+14(5.1e+13)- & \cellcolor[rgb]{ .851  .851  .851}4.40e+14(4.0e+13) & 0.609(6.5e-02)$\approx$ & \cellcolor[rgb]{ .851  .851  .851}0.597(5.9e-02) & 0.547(0.38)$\approx$ & \cellcolor[rgb]{ .851  .851  .851}0.578(0.37) \\
    German & 5.74e-04(2.2e-04)- & \cellcolor[rgb]{ .851  .851  .851}5.24e-04(2.1e-04) & \cellcolor[rgb]{ .851  .851  .851}4.05e+14(1.0e+14)$\approx$ & 3.92e+14(1.1e+14) & \cellcolor[rgb]{ .851  .851  .851}0.700(0.11)$\approx$ & 0.719(0.14) & 39.4(29)- & \cellcolor[rgb]{ .851  .851  .851}44.2(30) \\
    Student performance & \cellcolor[rgb]{ .851  .851  .851}4.66e-04(1.0e-04)+ & 5.04e-04(1.6e-04) & 3.78e+14(5.1e+13)- & \cellcolor[rgb]{ .851  .851  .851}4.06e+14(9.3e+13) & \cellcolor[rgb]{ .851  .851  .851}0.656(0.11)+ & 0.702(0.13) & 39.5(19)$\approx$ & \cellcolor[rgb]{ .851  .851  .851}42.0(23) \\
    COMPAS & \cellcolor[rgb]{ .851  .851  .851}1.90e-04(5.0e-05)$\approx$ & 2.00e-04(6.6e-05) & 3.76e+14(4.8e+13)- & \cellcolor[rgb]{ .851  .851  .851}4.07e+14(4.6e+13) & 0.569(7.9e-02)- & \cellcolor[rgb]{ .851  .851  .851}0.526(8.0e-02) & \cellcolor[rgb]{ .851  .851  .851}22.1(9.0)$\approx$ & 21.4(8.6) \\
    Bank  & 1.03e-04(6.6e-05)- & \cellcolor[rgb]{ .851  .851  .851}8.02e-05(7.9e-05) & \cellcolor[rgb]{ .851  .851  .851}4.2e+14(5.4e+13)+ & 4.15e+14(6.4e+13) & \cellcolor[rgb]{ .851  .851  .851}0.500(7.2e-02)+ & 0.528(7.4e-02) & 19.6(5.2)- & \cellcolor[rgb]{ .851  .851  .851}20.5(5.8) \\
    Adult & 7.30e-05(1.3e-05)- & \cellcolor[rgb]{ .851  .851  .851}5.03e-05(3.1e-05) & 4.9e+14(1.9e+13)- & \cellcolor[rgb]{ .851  .851  .851}5.34e+14(1.9e+13) & 0.574(5.0e-02)- & \cellcolor[rgb]{ .851  .851  .851}0.500(5.2e-02) & \cellcolor[rgb]{ .851  .851  .851}5.66(1.0)$\approx$ & 5.48(1.1) \\
    Drug consumption & 2.13e-04(1.1e-04)- & \cellcolor[rgb]{ .851  .851  .851}1.54e-04(8.6e-05) & 2.99e+14(3.5e+13)- & \cellcolor[rgb]{ .851  .851  .851}3.76e+14(3.8e+13) & \cellcolor[rgb]{ .851  .851  .851}0.485(6.8e-02)$\approx$ & 0.486(6.1e-02) & 7.45(4.7)- & \cellcolor[rgb]{ .851  .851  .851}12.9(5.7) \\
    Patient treatment & \cellcolor[rgb]{ .851  .851  .851}2.68e-04(7.8e-05)+ & 2.77e-04(1.2e-04) & 2.83e+14(8.9e+13)- & \cellcolor[rgb]{ .851  .851  .851}3.51e+14(1.1e+14) & \cellcolor[rgb]{ .851  .851  .851}0.640(0.10)$\approx$ & 0.663(0.17) & 40.8(20)- & \cellcolor[rgb]{ .851  .851  .851}47.3(15) \\
    LSAT  & 2.32e-04(1.0e-04)- & \cellcolor[rgb]{ .851  .851  .851}1.80e-04(7.3e-05) & 2.30e+14(4.2e+13)$\approx$ & \cellcolor[rgb]{ .851  .851  .851}2.36e+14(6.7e+13) & \cellcolor[rgb]{ .851  .851  .851}0.412(8.6e-02)+ & 0.456(7.7e-02) & 61.6(11)- & \cellcolor[rgb]{ .851  .851  .851}75.6(11) \\
    Default & 7.22e-05(2.4e-05)- & \cellcolor[rgb]{ .851  .851  .851}5.78e-05(3.0e-05) & 3.03e+14(1.8e+13)$\approx$ & \cellcolor[rgb]{ .851  .851  .851}3.04e+14(1.7e+13) & \cellcolor[rgb]{ .851  .851  .851}0.411(5.5e-02)$\approx$ & 0.412(5.1e-02) & 24.9(2.7)- & \cellcolor[rgb]{ .851  .851  .851}25.7(2.5) \\
    Dutch & 6.53e-05(6.0e-06)- & \cellcolor[rgb]{ .851  .851  .851}3.85e-05(1.4e-05) & 5.33e+14(2.1e+13)- & \cellcolor[rgb]{ .851  .851  .851}5.76e+14(2.3e+13) & 0.484(3.8e-02)$\approx$ & \cellcolor[rgb]{ .851  .851  .851}0.463(5.3e-02) & \cellcolor[rgb]{ .851  .851  .851}4.82(0.55)+ & 4.37(0.66) \\
    \midrule
      +/$\approx$/-    &   3/1/8    &   -    &   1/3/8  & -  &  3/6/3 & - & 1/4/7& -\\
    \bottomrule
    \end{tabular}%
    
  \label{tab:four_indicator2}%
\end{table*}%

\begin{table*}[htbp]
  \centering
  \caption{GD, PD, SP and HV values of final model set averaged over 50 trials. ``+/$\approx$/-'' indicates that the average indicator value of the corresponding algorithm (specified by column header) is statistically better/similar/worse than the one of $FaMOEL$ according to the Friedman test with a 0.05 significance level. The best averaged values are highlighted in grey.}
  \setlength{\tabcolsep}{2.5pt}
    \begin{tabular}{c|cc|cc|cc|cc}
    \toprule
       \multirow{2}{*}{Dataset}   & \multicolumn{2}{c|}{\textbf{GD}} & \multicolumn{2}{c|}{\textbf{PD}} & \multicolumn{2}{c|}{\textbf{SP}} & \multicolumn{2}{c}{\textbf{HV}}\\
     & $FaMOEL^-$ & $FaMOEL$  & $FaMOEL^-$ & $FaMOEL$ & $FaMOEL^-$ & $FaMOEL$  & $FaMOEL^-$ & $FaMOEL$\\
    \midrule
   Heart health & \cellcolor[rgb]{ .851  .851  .851}2.60e-(1.0e-04)+ & 2.88e-(8.3e-05) & 8.54e+(4.7e+13)- & \cellcolor[rgb]{ .851  .851  .851}2.08e+(4.5e+13) & \cellcolor[rgb]{ .851  .851  .851}0.537(0.11)+ & 0.660(9.8e-02) & 10.6(7.7)- & \cellcolor[rgb]{ .851  .851  .851}15.1(8.3) \\
    Titanic & 1.32e-(7.8e-05)$\approx$ & \cellcolor[rgb]{ .851  .851  .851}1.25e-(8.8e-05) & 2.47e+(1.1e+14)- & \cellcolor[rgb]{ .851  .851  .851}4.40e+(4.0e+13) & \cellcolor[rgb]{ .851  .851  .851}0.534(0.14)+ & 0.597(5.9e-02) & 0.316(0.21)- & \cellcolor[rgb]{ .851  .851  .851}0.578(0.37) \\
    German & \cellcolor[rgb]{ .851  .851  .851}3.95e-(1.7e-04)+ & 5.24e-(2.1e-04) & 2.55e+(1.8e+14)- & \cellcolor[rgb]{ .851  .851  .851}3.92e+(1.1e+14) & \cellcolor[rgb]{ .851  .851  .851}0.632(0.25)+ & 0.719(0.14) & 42.0(30)$\approx$ & \cellcolor[rgb]{ .851  .851  .851}44.2(30) \\    Student performance & \cellcolor[rgb]{ .851  .851  .851}4.10e-(1.0e-04)+ & 5.04e-(1.5e-04) & 3.14e+(8.5e+13)- & \cellcolor[rgb]{ .851  .851  .851}4.06e+(9.3e+13) & \cellcolor[rgb]{ .851  .851  .851}0.614(0.12)+ & 0.702(0.13) & \cellcolor[rgb]{ .851  .851  .851}46.1(22)+ & 42.0(23) \\
    COMPAS & \cellcolor[rgb]{ .851  .851  .851}1.48e-(5.6e-05)+ & 2.00e-(6.6e-05) & 2.45e+(9.1e+13)- & \cellcolor[rgb]{ .851  .851  .851}4.07e+(4.6e+13) & \cellcolor[rgb]{ .851  .851  .851}0.390(0.11)+ & 0.526(8.0e-02) & 20.9(9.4)$\approx$ & \cellcolor[rgb]{ .851  .851  .851}21.4(8.6) \\
    Bank  & 8.88e-(5.5e-05)- & \cellcolor[rgb]{ .851  .851  .851}8.02e-(7.8e-05) & 3.35e+(5.0e+13)- & \cellcolor[rgb]{ .851  .851  .851}4.15e+(6.4e+13) & \cellcolor[rgb]{ .851  .851  .851}0.438(6.7e-02)+ & 0.528(7.4e-02) & 20.4(5.9)$\approx$ & \cellcolor[rgb]{ .851  .851  .851}20.5(5.8) \\
    Adult & 6.89e-(1.4e-05)- & \cellcolor[rgb]{ .851  .851  .851}5.03e-(3.1e-05) & 4.89e+(3.9e+13)- & \cellcolor[rgb]{ .851  .851  .851}5.34e+(1.9e+13) & \cellcolor[rgb]{ .851  .851  .851}0.435(5.0e-02)+ & 0.500(5.2e-02) & 4.66(1.2)- & \cellcolor[rgb]{ .851  .851  .851}5.48(1.1) \\
    Drug consumption & \cellcolor[rgb]{ .851  .851  .851}1.35e-(5.4e-05)+ & 1.54e-(8.5e-05) & 2.09e+(9.8e+13)- & \cellcolor[rgb]{ .851  .851  .851}3.76e+(3.8e+13) & \cellcolor[rgb]{ .851  .851  .851}0.369(0.13)+ & 0.486(6.1e-02) & 11.3(6.0)- & \cellcolor[rgb]{ .851  .851  .851}12.9(5.7) \\
    Patient treatment & \cellcolor[rgb]{ .851  .851  .851}1.59e-(7.9e-05)+ & 2.77e-(1.2e-04) & 1.70e+(9.8e+13)- & \cellcolor[rgb]{ .851  .851  .851}3.51e+(1.1e+14) & \cellcolor[rgb]{ .851  .851  .851}0.345(0.19)+ & 0.663(0.17) & 42.1(20)- & \cellcolor[rgb]{ .851  .851  .851}47.3(15) \\
    LSAT  & \cellcolor[rgb]{ .851  .851  .851}1.14e-(3.1e-05)+ & 1.80e-(7.2e-05) & 1.37e+(6.6e+13)- & \cellcolor[rgb]{ .851  .851  .851}2.36e+(6.7e+13) & \cellcolor[rgb]{ .851  .851  .851}0.333(0.11)+ & 0.456(7.7e-02) & 75.3(12)$\approx$ & \cellcolor[rgb]{ .851  .851  .851}75.6(11) \\
    Default & \cellcolor[rgb]{ .851  .851  .851}5.28e-(1.8e-05)$\approx$ & 5.78e-(2.9e-05) & 2.37e+(3.3e+13)- & \cellcolor[rgb]{ .851  .851  .851}3.04e+(1.7e+13) & \cellcolor[rgb]{ .851  .851  .851}0.357(6.9e-02)+ & 0.412(5.1e-02) & 25.3(2.7)$\approx$ & \cellcolor[rgb]{ .851  .851  .851}25.7(2.5) \\
    Dutch & 5.45e-(5.1e-06)- & \cellcolor[rgb]{ .851  .851  .851}3.85e-(1.4e-05) & 4.19e+(4.4e+13)- & \cellcolor[rgb]{ .851  .851  .851}5.76e+(2.3e+13) & \cellcolor[rgb]{ .851  .851  .851}0.369(5.1e-02)+ & 0.463(5.3e-02) & \cellcolor[rgb]{ .851  .851  .851}4.88(1.1)+ & 4.37(0.66) \\
    \midrule
    +/$\approx$/- & 7/2/3  &  -  & 0/0/12 &  -  & 12/0/0  &  -  & 2/5/5 &  -  \\
    \bottomrule
    \end{tabular}%
  \label{tab:four_indicator3}%
\end{table*}%

\def\removed{
\begin{table}[htbp]
  \centering
  \caption{GD values of final model set averaged over 50 trials. ``+/$\approx$/-'' indicates that the average GD value of the corresponding algorithm (specified by column header) is statistically better/similar/worse than the one of $FaMOEL$ according to the Friedman test with a 0.05 significance level. The best averaged GD values are highlighted in grey backgrounds.}
  \begin{adjustbox} {max width=0.45\textwidth}
    \begin{tabular}{ccc}
    \toprule
    Dataset & $MOEL_{Rep}^-$ & $FaMOEL$ \\
    \midrule
    Heart health & \cellcolor[rgb]{ .851  .851  .851 }2.6667e-04(9.299e-05)+ & 2.8849e-04(8.331e-05) \\
   Titanic & 1.3908e-04(7.737e-05)- & \cellcolor[rgb]{ .851  .851  .851 }1.2584e-04(8.818e-05) \\
    German & 5.7440e-04(2.183e-04)- & \cellcolor[rgb]{ .851  .851  .851 }5.2444e-04(2.119e-04) \\
    Student performance & \cellcolor[rgb]{ .851  .851  .851 }4.6665e-04(1.077e-04)+ & 5.0436e-04(1.592e-04) \\
    COMPAS & \cellcolor[rgb]{ .851  .851  .851 }1.9014e-04(4.975e-05)$\approx$ & 2.0006e-04(6.629e-05) \\
    Bank & 1.0374e-04(6.556e-05)- & \cellcolor[rgb]{ .851  .851  .851 }8.0269e-05(7.885e-05) \\
    Adult & 7.3056e-05(1.339e-05)- & \cellcolor[rgb]{ .851  .851  .851 }5.0351e-05(3.165e-05) \\
    Drug consumption & 2.1332e-04(1.096e-04)- & \cellcolor[rgb]{ .851  .851  .851 }1.5466e-04(8.551e-05) \\
    Patient treatment & \cellcolor[rgb]{ .851  .851  .851 }2.6810e-04(7.770e-05)+ & 2.7786e-04(1.210e-04) \\
    LSAT & 2.3262e-04(1.077e-04)- & \cellcolor[rgb]{ .851  .851  .851 }1.8065e-04(7.299e-05) \\
    Default & 7.2291e-05(2.412e-05)- & \cellcolor[rgb]{ .851  .851  .851 }5.7849e-05(2.992e-05) \\
    Dutch & 6.5343e-05(5.957e-06)- & \cellcolor[rgb]{ .851  .851  .851 }3.8519e-05(1.422e-05) \\
    \midrule
     +/$\approx$/- & 3/1/8 &  -   \\    
    \bottomrule
    \end{tabular}%
    \end{adjustbox}
  \label{tab:GD2}%
\end{table}%

\begin{table}[htbp]
  \centering
  \caption{PD values of final model set averaged over 50 trials. ``+/$\approx$/-'' indicates that the average PD value of the corresponding algorithm (specified by column header) is statistically better/similar/worse than the one of $FaMOEL$ according to the Friedman test with a 0.05 significance level. The best averaged PD values are highlighted in grey backgrounds.}
  \begin{adjustbox} {max width=0.45\textwidth}
    \begin{tabular}{ccc}
    \toprule
    Dataset & $MOEL_{Rep}^-$ & $FaMOEL$ \\
    \midrule
    Heart health & 1.4216e+14(3.114e+13)- & \cellcolor[rgb]{ .851  .851  .851 }2.0844e+14(4.550e+13) \\
   Titanic & 4.0475e+14(5.144e+13)- & \cellcolor[rgb]{ .851  .851  .851 }4.4054e+14(4.023e+13) \\
    German & \cellcolor[rgb]{ .851  .851  .851 }4.0538e+14(1.094e+14)$\approx$ & 3.9227e+14(1.143e+14) \\
    Student performance & 3.7836e+14(5.175e+13)- & \cellcolor[rgb]{ .851  .851  .851 }4.0686e+14(9.340e+13) \\
    COMPAS & 3.7635e+14(4.887e+13)- & \cellcolor[rgb]{ .851  .851  .851 }4.0793e+14(4.670e+13) \\
    Bank & \cellcolor[rgb]{ .851  .851  .851 }4.279e+14(5.424e+13)+ & 4.1577e+14(6.493e+13) \\
    Adult & 4.981e+14(1.942e+13)- & \cellcolor[rgb]{ .851  .851  .851 }5.3427e+14(1.957e+13) \\
    Drug consumption & 2.9924e+14(3.587e+13)- & \cellcolor[rgb]{ .851  .851  .851 }3.7684e+14(3.835e+13) \\
    Patient treatment & 2.8334e+14(8.949e+13)- & \cellcolor[rgb]{ .851  .851  .851 }3.5135e+14(1.148e+14) \\
    LSAT & 2.3042e+14(4.264e+13)$\approx$ & \cellcolor[rgb]{ .851  .851  .851 }2.3627e+14(6.704e+13) \\
    Default & 3.0314e+14(1.807e+13)$\approx$ & \cellcolor[rgb]{ .851  .851  .851 }3.0483e+14(1.722e+13) \\
    Dutch & 5.3393e+14(2.141e+13)- & \cellcolor[rgb]{ .851  .851  .851 }5.7663e+14(2.322e+13) \\
    \midrule
     +/$\approx$/- & 1/3/8 &  -   \\    
    \bottomrule
    \end{tabular}%
    \end{adjustbox}
  \label{tab:PD2}%
\end{table}%

\begin{table}[htbp]
  \centering
  \caption{SP values of final model set averaged over 50 trials. ``+/$\approx$/-'' indicates that the average SP value of the corresponding algorithm (specified by column header) is statistically better/similar/worse than the one of $FaMOEL$ according to the Friedman test with a 0.05 significance level. The best averaged SP values are highlighted in grey backgrounds.}
  \begin{adjustbox} {max width=0.45\textwidth}
    \begin{tabular}{ccc}
    \toprule
    Dataset & $MOEL_{Rep}^-$ & $FaMOEL$ \\
    \midrule
     Heart health & 7.1059e-01(9.665e-02)- & \cellcolor[rgb]{ .851  .851  .851 }6.6010e-01(9.849e-02) \\
   Titanic & 6.0960e-01(6.515e-02)$\approx$ & \cellcolor[rgb]{ .851  .851  .851 }5.9740e-01(5.970e-02) \\
    German & \cellcolor[rgb]{ .851  .851  .851 }7.0062e-01(1.131e-01)$\approx$ & 7.1968e-01(1.353e-01) \\
    Student performance & \cellcolor[rgb]{ .851  .851  .851 }6.5660e-01(1.137e-01)+ & 7.0276e-01(1.302e-01) \\
    COMPAS & 5.6922e-01(7.944e-02)- & \cellcolor[rgb]{ .851  .851  .851 }5.2633e-01(8.019e-02) \\
    Bank & \cellcolor[rgb]{ .851  .851  .851 }5.0048e-01(7.233e-02)+ & 5.2882e-01(7.451e-02) \\
    Adult & 5.7495e-01(5.085e-02)- & \cellcolor[rgb]{ .851  .851  .851 }5.0053e-01(5.264e-02) \\
    Drug consumption & \cellcolor[rgb]{ .851  .851  .851 }4.8544e-01(6.847e-02)$\approx$ & 4.8652e-01(6.127e-02) \\
    Patient treatment & \cellcolor[rgb]{ .851  .851  .851 }6.4084e-01(1.044e-01)$\approx$ & 6.6337e-01(1.685e-01) \\
    LSAT & \cellcolor[rgb]{ .851  .851  .851 }4.1240e-01(8.649e-02)+ & 4.5648e-01(7.716e-02) \\
    Default & \cellcolor[rgb]{ .851  .851  .851 }4.1148e-01(5.591e-02)$\approx$ & 4.1220e-01(5.199e-02) \\
    Dutch & 4.8456e-01(3.875e-02)$\approx$ & \cellcolor[rgb]{ .851  .851  .851 }4.6399e-01(5.316e-02) \\
    \midrule
     +/$\approx$/- & 3/6/3 &  -   \\    
    \bottomrule
    \end{tabular}%
    \end{adjustbox}
  \label{tab:SP2}%
\end{table}%

\begin{table}[htbp]
  \centering
  \caption{HV values of final model set averaged over 50 trials. ``+/$\approx$/-'' indicates that the average HV value of the corresponding algorithm (specified by column header) is statistically better/similar/worse than the one of $FaMOEL$ according to the Friedman test with a 0.05 significance level. The best averaged HV values are highlighted in grey backgrounds.}
  \begin{adjustbox} {max width=0.45\textwidth}
    \begin{tabular}{ccc}
    \toprule
    Dataset & $MOEL_{Rep}^-$ & $FaMOEL$ \\
    \midrule
    Heart health & 1.0047e+01(7.727e+00)- & \cellcolor[rgb]{ .851  .851  .851 }1.5164e+01(8.302e+00) \\
   Titanic & 5.4737e-01(3.801e-01)$\approx$ & \cellcolor[rgb]{ .851  .851  .851 }5.7825e-01(3.723e-01) \\
    German & 3.9459e+01(2.845e+01)- & \cellcolor[rgb]{ .851  .851  .851 }4.4298e+01(2.991e+01) \\
    Student performance & 3.9597e+01(1.908e+01)$\approx$ & \cellcolor[rgb]{ .851  .851  .851 }4.2085e+01(2.309e+01) \\
    COMPAS & \cellcolor[rgb]{ .851  .851  .851 }2.2120e+01(8.977e+00)$\approx$ & 2.1401e+01(8.632e+00) \\
    Bank & 1.9689e+01(5.186e+00)- & \cellcolor[rgb]{ .851  .851  .851 }2.0535e+01(5.819e+00) \\
    Adult & \cellcolor[rgb]{ .851  .851  .851 }5.6602e+00(1.037e+00)$\approx$ & 5.4880e+00(1.071e+00) \\
    Drug consumption & 7.4510e+00(4.666e+00)- & \cellcolor[rgb]{ .851  .851  .851 }1.2918e+01(5.665e+00) \\
    Patient treatment & 4.0862e+01(1.936e+01)- & \cellcolor[rgb]{ .851  .851  .851 }4.7353e+01(1.477e+01) \\
    LSAT & 6.1664e+01(1.072e+01)- & \cellcolor[rgb]{ .851  .851  .851 }7.5694e+01(1.138e+01) \\
    Default & 2.4904e+01(2.700e+00)- & \cellcolor[rgb]{ .851  .851  .851 }2.5774e+01(2.465e+00) \\
    Dutch & \cellcolor[rgb]{ .851  .851  .851 }4.8290e+00(5.463e-01)+ & 4.3745e+00(6.545e-01) \\
    \midrule
     +/$\approx$/- & 1/4/7 &  -   \\    
    \bottomrule
    \end{tabular}%
    \end{adjustbox}
  \label{tab:HV2}%
\end{table}%
}

\subsection{Effectiveness of Our Fairness-aware Strategy}\label{sec:fairness_aware_sensitivity}

Our fairness-aware enhancement strategy, as described in Section \ref{sec:Fairness_aware_strategy}, is designed to improve the work~\cite{wang2016objective}. In order to evaluate the effectiveness of our enhancement, we compare our approach $FaMOEL$ with the method that utilises the original strategy proposed~\cite{wang2016objective}, denoted as $FaMOEL^-$. The difference between $FaMOEL$ and $FaMOEL^-$ was discussed in Section \ref{sec:Fairness_aware_strategy}. $FaMOEL^-$ is implemented with the same settings as ones of $FaMOEL$ except for the fairness-aware strategy. 

Table \ref{tab:four_indicator3} presents the results of GD, PD, SP and HV obtained by $FaMOEL^-$ and $FaMOEL$, respectively. $FaMOEL$ outperforms $FaMOEL^-$ and achieves no worse performance on 10 out of 12 datasets. $FaMOEL^-$ demonstrates better convergence and uniformity performance as measured by GD and SP, respectively. However, $FaMOEL^-$ exhibits weaker spread performance. This suggests that the model set obtained by $FaMOEL^-$ only converges to a subregion, resulting in a significant loss in overall performance, despite being closer to the Pareto front compared to $FaMOEL$. Thus, our enhancement strategies are specifically designed to prevent the model set from getting trapped in local regions, such as designing a warm starting and constructing a robust mNCIE matrix. Also, the selection threshold $\tau$ is to avoid deleting objectives that are weakly positively correlated with $\mathcal{E}_J$ (line \ref{line:cut} in Algorithm \ref{algo:reduction}). This strategy ensures that $\mathcal{E}'$ can adequately represent the entire $\mathcal{E}$.

In Student performance and Dutch, $FaMOEL$ performs worse than $FaMOEL^-$ in terms of HV. 
In Student performance, compared with $FaMOEL^-$, $FaMOEL$ exhibits better PD performance but worse GD and SP performance. This suggests that the model set obtained by $FaMOEL$ is situated too far away from the Pareto front and loses a significant overall performance. 
As for Dutch, $FaMOEL$ has better GD and PD but worse SP performance. One potential explanation for the inferior HV performance of $FaMOEL$ is that the model set obtained by $FaMOEL$ has poorer uniformity, which ultimately leads to a substantial degradation in overall performance.

In summary, our proposed fairness-aware enhancement strategy can contribute to constructing a more suitable representative subset of all the considered objectives to guide the model training process, especially in improving the spread performance of the learning models.

\subsection{Computational Cost Analysis}
\begin{figure}[bh]
  \begin{center} 
      \includegraphics[width=0.48\textwidth]{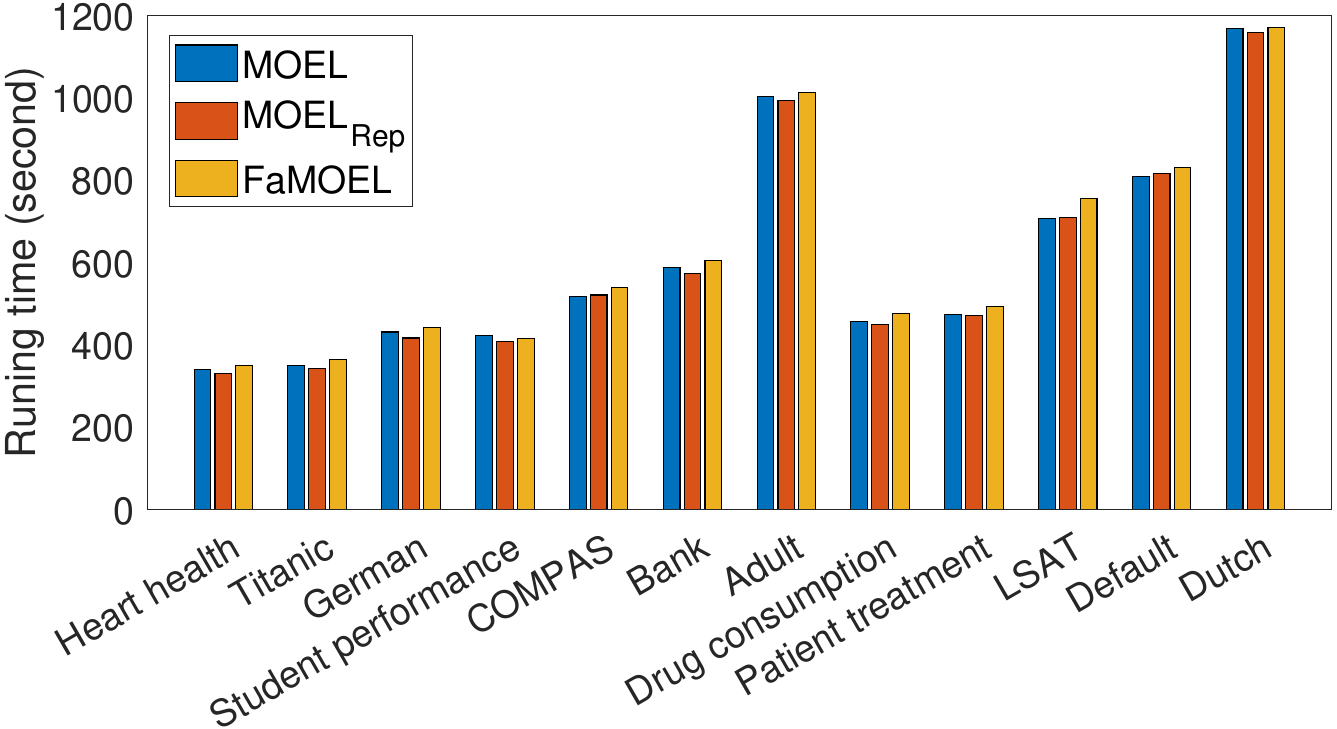}
  \end{center}
  \caption{Computational time cost of $MOEL$, $MOEL_{Rep}$ and $FaMOEL$ on 12 datasets.
    } \label{runtime}
\end{figure}

To analyse the efficiency of $FaMOEL$, we report the average runtime of $MOEL$, $MOEL_{Rep}$, and $FaMOEL$ in Fig. 7. Overall, $FaMOEL$ have a similar computation runtime to $MOEL$ and $MOEL_{Rep}$, as indicated in Fig.~\ref{runtime}.
Nonetheless, it's worth noting that $MOEL_{Rep}$ which involves two issues relies on a set of pre-defined fairness measures. First, determining suitable measures requires considerable computational cost as different algorithms should be run across various datasets to identify the correlation among the measures. Secondly, those pre-defined measures may show different correlation on a new dataset. Therefore, the results demonstrate the effectiveness of our framework.

\subsection{Parameter Sensitivity Analysis}\label{sec:Parameter_sensitivity}

In this study, we aim to analyse the sensitivity of the unique hyperparameter in our algorithms $\tau$ and recommend a value. The parameter $\tau$ is introduced in our enhanced fairness-aware strategy and serves as a selection threshold for determining the objective set $Del$ to be removed. Each objective in $Del$ is viewed to be highly positively correlated with the most representative objective $\mathcal{E}_J$ in the objective set indexed by $S$ (cf. Algorithm \ref{algo:reduction}). The sensitivity analysis of $\tau$ involves two steps: (i) applying a set of coarse-grained $\tau$ values $\{0.1, 0.2, 0.3, 0.4, 0.5\}$ to all 12 datasets, (ii) selecting two datasets that exhibit higher sensitivity to $\tau$ and conducting a fine-grained analysis using a set of $\tau$ values to these two datasets.

\begin{figure*}[!htb]
  \begin{center}  \includegraphics[width=1\textwidth]{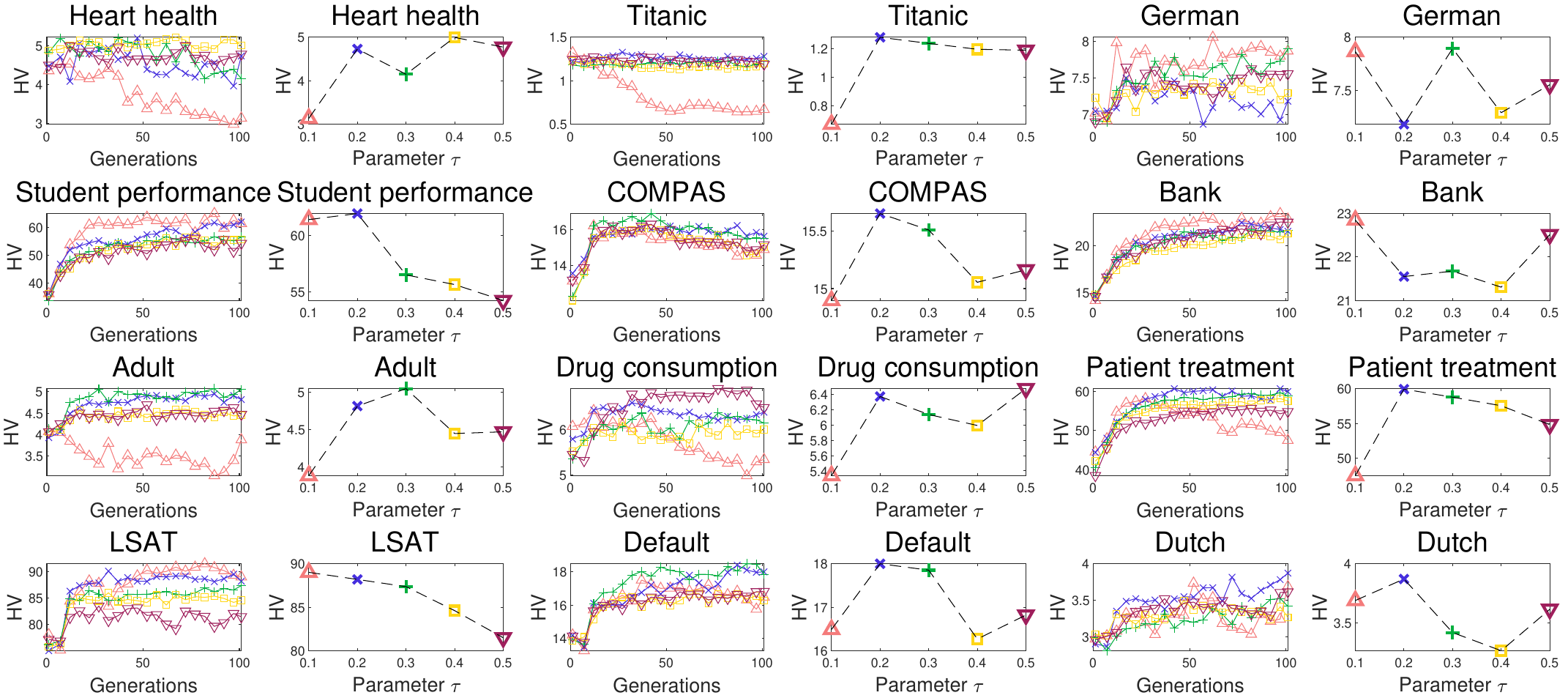}
  \end{center}
  \caption{Averaged HV values of $FaMOEL$ under coarse-grained $\tau$ values $\{0.1, 0.2, 0.3, 0.4, 0.5\}$. For each dataset, the left figure is for the averaged HV values along with generations. The right figure is for the averaged HV values in the final generation along with different parameters $\tau$.
  } \label{fig:para_sensitive1}
\end{figure*}

\begin{figure}[htbp]
  \begin{center}  \includegraphics[width=0.5\textwidth]{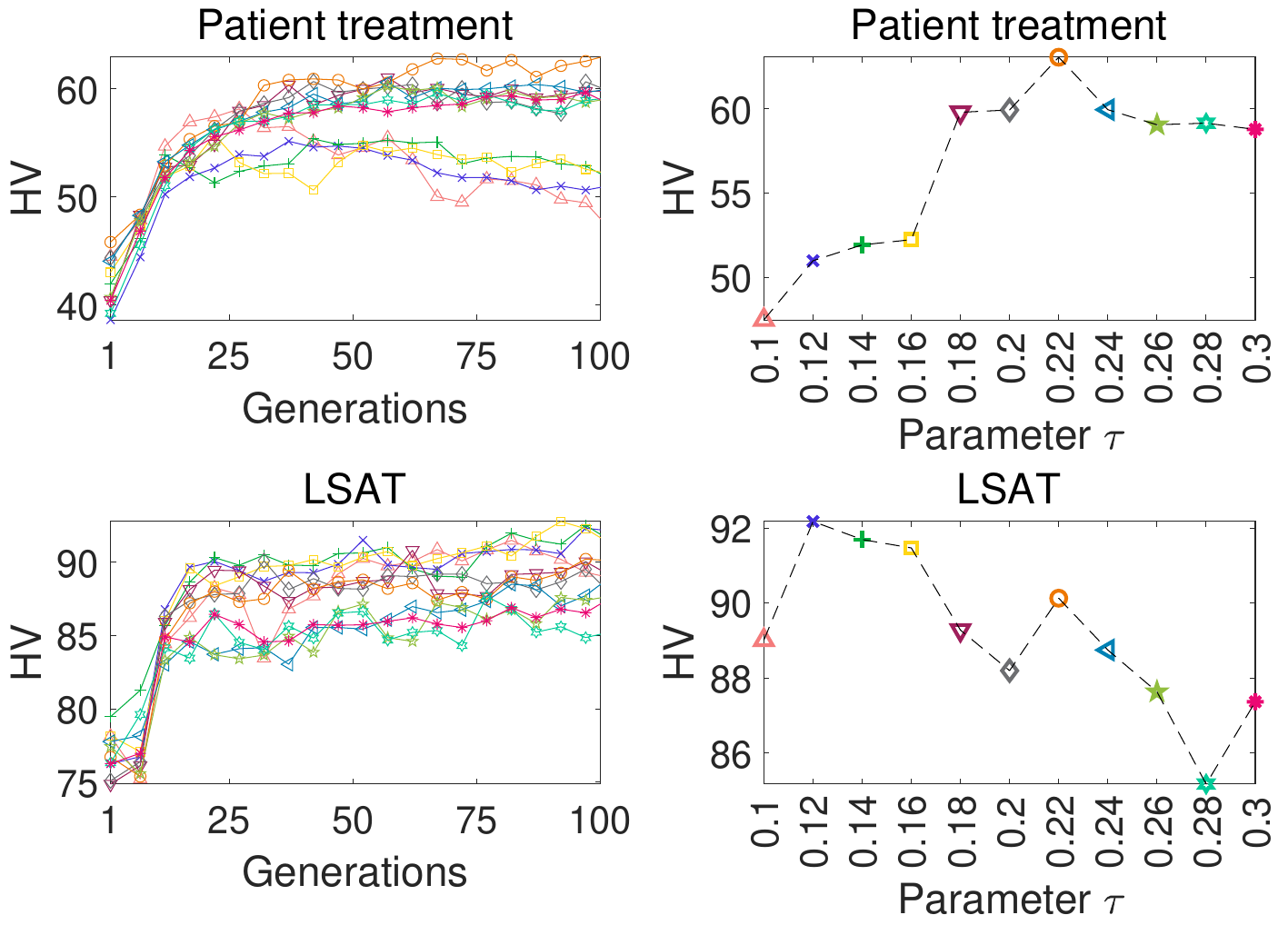}
  \end{center}
  \caption{Averaged HV values of $FaMOEL$ under fine-grained $\tau$ values ranging from $0.1$ to $0.3$ with step 0.02. For each dataset, the left figure is for the averaged HV values along with generations. The right figure is for the averaged HV values in the final generation along with different parameters $\tau$.
  } \label{fig:para_sensitive2}
\end{figure}

Fig. \ref{fig:para_sensitive1} presents how the HV performance of $FaMOEL$ varies with the $\tau$ set $\{0.1, 0.2, 0.3, 0.4, 0.5\}$. The influences of $\tau$ exhibit diverse patterns across 12 datasets but have a similar observation. Considering all the 12 datasets, except for Heart health and Drug consumption, the better performance of HV falls into the interval $[0.1, 0.3]$. In addition, it also demonstrates that both Patient treatment and LSAT show a higher sensitivity to $\tau$. To further investigate the influence of $\tau$ in a more detailed manner, we conduct a fine-grained analysis using a set of $\tau$ values ranging from $0.1$ to $0.3$ with step 0.02, as depicted in Fig. \ref{fig:para_sensitive2}. Based on the results from these two datasets, a selection threshold value of $0.22$ appears to be a preferable choice for $\tau$.   In general, $\tau$ is a problem dependent hyper-parameter.

\section{Conclusion}\label{sec:conclusion}

When considering a set of fairness measures, this paper proposes to dynamically and adaptively determine a representative subset of measures as optimisation objectives during model training without relying on any prior knowledge. The determined set can be used as objectives of multiobjective evolutionary learning to guide the evolution of learning models.
Extensive experimental studies demonstrate that our framework achieves very good performance in dealing with accuracy and 25 fairness measures. 
Furthermore, it is observed that the selection of suitable objectives varies across different training stages, which our fairness-aware strategy effectively detects. Compared with the state-of-the-art algorithm optimising a static representative subset, our method eliminates the need for prior knowledge in determining the representative subset and achieves superior performance in general. It is also worth noting that our work represents one of the few attempts in machine learning where the learning objectives (or loss functions) change adaptively during training.

In the future, we plan to explore our work. 
As shown in Fig. \ref{fig:para_sensitive1}, the optimal parameter $\tau$ varies across different datasets. Therefore, an adaptive mechanism for tuning the parameter $\tau$ is required to determine a more appropriate subset of representative measures along with model training. We can also employ one of the existing methods for tuning parameters automatically~\cite{8733017}. 
Furthermore, we plan to enhance the feasibility and interoperability of our framework when applied to more complex models, such as deep learning models.

\balance
\bibliographystyle{IEEEtran}
\bibliography{fairness}

\end{document}